\documentclass[a4paper,fleqn]{cas-dc}

\usepackage[numbers]{natbib}

\def\tsc#1{\csdef{#1}{\textsc{\lowercase{#1}}\xspace}}
\tsc{WGM}
\tsc{QE}
\tsc{EP}
\tsc{PMS}
\tsc{BEC}
\tsc{DE}

\begin{document}
\let\WriteBookmarks\relax
\def\floatpagepagefraction{1}
\def\textpagefraction{.001}
\shorttitle{Bayesian Inference Neural Surrogate for SPN models}
\shortauthors{B.K. Manu et~al.}

\title [mode = title]{A Simple Approximate Bayesian Inference Neural Surrogate for Stochastic Petri Net Models}                      



\author[1]{Bright Kwaku Manu}[type=editor,
                        orcid=0009-0004-0805-0915]
\ead{bkmanu@asu.edu}


\affiliation[1]{organization={School of Computing and Augmented Intelligence, Arizona State University},
                postcodesep={},
                city={Tempe},
                postcode={85281}, 
                state={Arizona},
                country={USA}}

\author[2]{Trevor Reckell}
\ead{trreckel@asu.edu}

\author[3]{Beckett Sterner}[%
   ]
\cormark[1]
\ead{beckett.sterner@asu.edu}


\affiliation[2]{organization={School of Mathematical and Statistical Sciences, Arizona State University},
                postcodesep={}, 
                city={Tempe},
                postcode={85281},
                state={Arizona},
                country={USA}}

\affiliation[3]{organization={School of Life Sciences, Arizona State University},
                postcodesep={},
                city={Tempe},
                postcode={85281}, 
                state={Arizona}, 
                country={USA}}

\cortext[cor1]{Corresponding author}


\affiliation[2]{organization={School of Mathematical and Statistical Sciences, Arizona State University},
                postcodesep={},
                city={Tempe},
                postcode={85281},
                state={Arizona},
                country={USA}}

\author[4]{Petar Jevti\'c}
\ead{petar.jevtic@asu.edu}

\begin{abstract}
Stochastic Petri Nets (SPNs) are an increasingly popular tool of choice for modeling discrete‐event dynamics in areas such as epidemiology and systems biology, yet their parameter estimation remains challenging in general and in particular when transition rates depend on external covariates and explicit likelihoods are unavailable. We introduce a neural-surrogate (neural-network-based approximation of the posterior distribution) framework that predicts the coefficients of known covariate-dependent rate functions directly from noisy, partially observed token trajectories. Our model employs a lightweight 1D Convolutional Residual Network trained end‐to‐end on Gillespie‐simulated SPN realizations, learning to invert system dynamics under realistic conditions of event dropout. During inference, Monte Carlo dropout provides calibrated uncertainty bounds together with point estimates. On synthetic SPNs with $\textbf{10\%}$ missing events, our surrogate recovers rate-function coefficients with an RMSE $= 0.043$ and substantially runs faster than traditional Bayesian approaches. These results demonstrate that data-driven, likelihood-free surrogates can enable accurate, robust, and real-time parameter recovery in complex, partially observed discrete-event systems.
\end{abstract}



\begin{keywords}
Petri Net \sep Stochastic Petri Net \sep Epidemiological Modeling \sep Bayesian Inference \sep Neural Network \sep Monte Carlo Dropout \par\noindent\rule{4.3cm}{0.5pt}\par
\textit{Code Link:}\sep \url{https://github.com/BrightManu-lang/SPN-param-recovery.git}
\end{keywords}

\maketitle

\section{Introduction}

Mathematical models play a major role in understanding and predicting the behavior of complex dynamical systems. Traditionally, ordinary differential equations (ODEs) have served as the workhorse for modeling continuous processes in fields such as systems biology, chemical kinetics, and epidemiology. Their ability to represent smooth temporal dynamics has enabled the analysis and control of many biological and engineered systems. However, ODEs are often inadequate for systems where the dynamics are inherently stochastic, discrete, or involve concurrent events. To address these limitations, Petri Nets have emerged as a powerful alternative modeling framework.

Petri Nets have found widespread use in epidemiology in capturing discrete and stochastic dynamics. Frameworks such as GSPN, colored Petri Nets, and geometric Petri Net approaches have been applied to COVID 19 and SEIR-type systems \cite{peng2021modeling, connolly2022epidemic, segovia2025petri, bahi2003modeling}. More recently, a compelling study has been done to demonstrate how Petri Nets approximate to an ODE solution under certain numerical conditions \cite{reckell2025numericalcomparisonpetrinet}. While parameter recovery for continuous dynamics and agent-based epidemiological models via likelihood-based, Approximate Bayesian Computation, or neural-ODE methods is now well-established \cite{yazdani2020systems, gaskin2023neural, radev2021outbreakflow}, discrete-event Stochastic Petri Nets (SPNs), which lack closed-form likelihoods, remain out of reach for these techniques.

To address this gap, we propose a neural surrogate framework that enables fast, likelihood-free recovery of transition rate parameters in SPNs, specifically targeting cases where transition rates are governed by known functional relationships with external covariates. Instead of estimating the raw transition rates, our model learns to predict the coefficients of these covariate-dependent functions directly from simulated Petri Net token trajectories. The surrogate is trained end-to-end on Gillespie-simulated SPN realizations and integrates 1D-convolutional residual network to extract temporal features. At inference time, we apply Monte Carlo dropout to quantify uncertainty, yielding calibrated confidence bounds alongside point estimates. This approach enables interpretable, data-driven adaptation of SPN behavior in response to changing environmental conditions, for example temperature, relative humidity, rainfall etc., an essential capability for real-time decision making in complex, partially observed systems.

The rest of the paper is organized as follows. Section II reviews the standard Petri Net formalism and their epidemiological applications, surveying key simulation-based inference techniques. Section III introduces notation, presents the SPN simulation model and the neural-surrogate training model, describes how functional transition rates are computed from environmental covariates, and details the underlying coefficient estimation procedure. Section IV details the training setup, and Section V presents the experimental findings. Finally, Section VI outlines directions for future work.

\section{Related Work}
\subsection{Petri Nets: An overview}
Petri Nets, introduced by Carl Petri\cite{petri1966communication}, are a foundational formalism for modeling and analyzing concurrent, distributed, and asynchronous systems including biological systems. Petri Net models consist of places, transitions and directed arcs. Places act as containers of tokens, and the distribution of tokens over places, commonly referred to as a marking, represents the current state of the system. Transitions model events in such a way that their firing changes the marking by consuming tokens from input places and producing tokens in output places \cite{petri1966communication}.

Mathematically \cite{peterson1977petri, sankaranarayanan2003petri, petri1966communication}, a Petri Net is a four-tuple $(\textbf{P}, \textbf{T}, \textbf{I}, \textbf{O})$ formalism together with an initial marking $m_0$, where $\textbf{P}$ denotes the places, $\textbf{T}$ the transitions, $\textbf{I}$ a multi-set of input places and $\textbf{O}$ a multi-set of output places. A Petri Net's state is given by a marking $m:\textbf{P}\rightarrow \mathbb{N}$, representing the number of tokens in each place. Given the initial conditions $m_0$, the net evolves by firing enabled transitions. A transition $t\in \mathbf{T}$ is enabled in marking $m$ if, for every place $p$, $m(p)\geq \textbf{I}(t)(p)$. Firing $t$ yields a new marking $m'$ such that $m'(p)=m(p)-\textbf{I}(t)+\textbf{O}(t)(p),$ ~~$\forall~ p\in \textbf{P}$.

\begin{figure}
    \centering
    \includegraphics[width=0.5\linewidth]{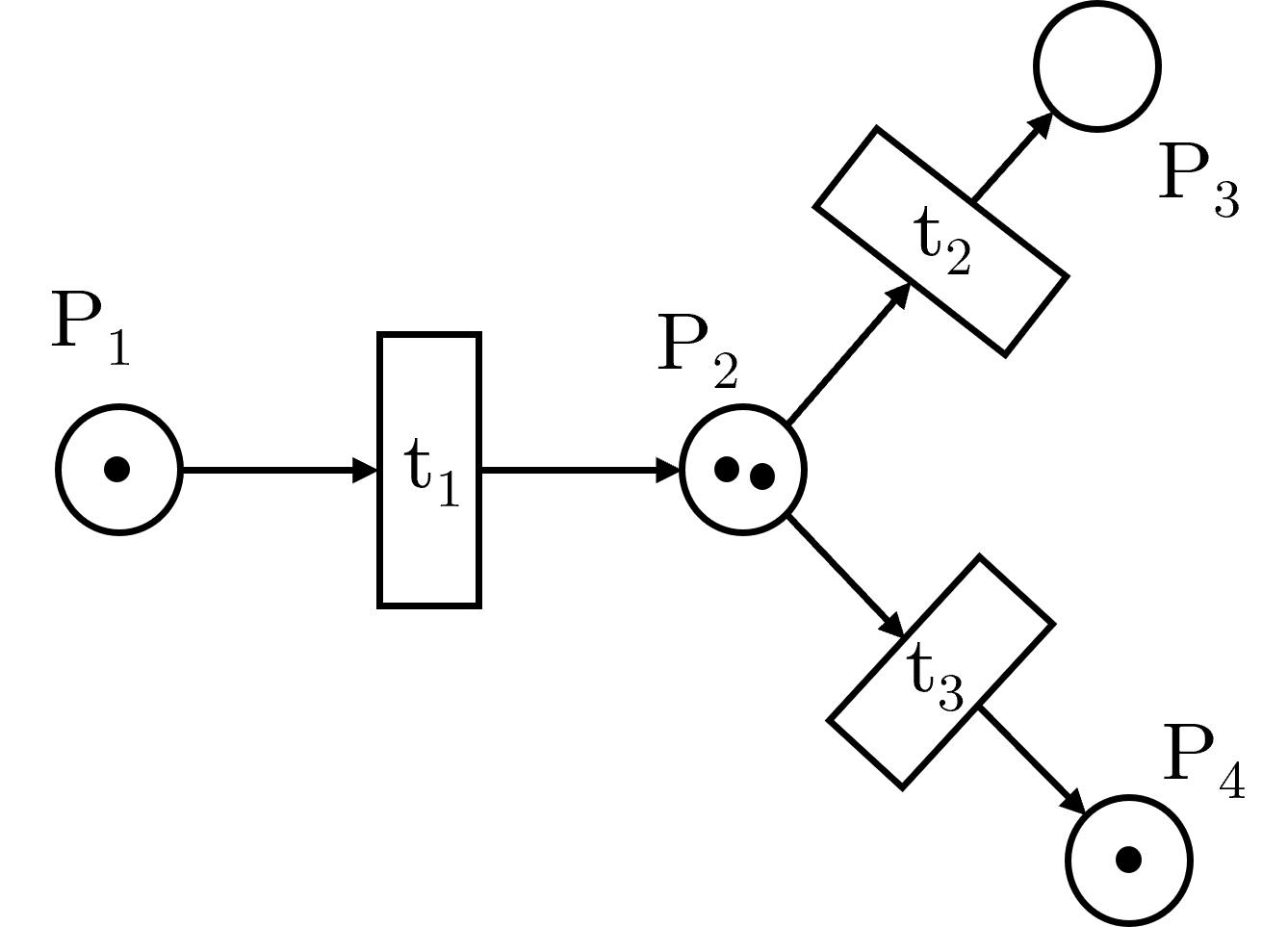}
    \caption{A visual representation of a simple Petri Net. $P_i's$ represent the places (circles) which contains the markings (dark dots). $t_i's$ denotes the transitions (rectangles). The arrows are directed arcs between a place and a transition, which can be assigned specific weights}
    \label{fig:enter-label}
\end{figure}

\subsection{Petri Nets in Epidemiology}
Recent applications of Petri Nets to epidemiological modeling have demonstrated its ability to integrate discrete events, continuous dynamics, and stochastic variability into a single graphical model.
For example, Peng et al. \cite{peng2021modeling} proposed a Generalized Stochastic Petri Net (GSPN) framework to model COVID-19 and other epidemic scenarios. They combine discrete-event concurrency with isomorphic Markov-chain analysis to evaluate both natural history and intervention strategies such as vaccination delays and reporting lags, at a system level. The authors in \cite{bahi2003modeling} introduced Colored Stochastic Petri Nets (CSPNs) as a unified graphical formalism for modeling directly transmitted infectious diseases by representing host heterogeneity and disease states within a single net structure. In their work, each structural place corresponds to an at-risk class of hosts, while token colors encodes serological states such as susceptible versus infected, and transitions, parameterized by arbitrary distributions, capture demographic, behavioral and epidemiological events such as recruitment, death and transmission. They demonstrated that density and frequency-dependent incidence naturally arise from encounter transitions with probabilistic color-change rules. Connolly et al. \cite{connolly2022epidemic} extended this idea with Colored Stochastic Petri Nets to build multilevel meta-population pandemic models that incorporate age and location stratification via travel graphs. They built a PetriNuts platform which supports deterministic, stochastic, and hybrid simulation including structural and behavioral analysis.

Two recent efforts have extended Petri Net modeling frameworks to incorporate the formal computation of the basic reproduction number $R_0$, a key quantity in infectious disease dynamics. Segovia \cite{segovia2025petri} proposed a geometric construction of the next generation matrix directly on the Petri Net structure. The approach decomposes the net into three functional subnets representing susceptible module, infection processes module, and infected module, and shows how the classical next generation approach by van den Driessche-Watmough maps onto these subnets. Under this formulation, the flow of infections through the net yields a matrix whose spectral radius gives the values of $R_0$. In parallel, Reckell et al. \cite{reckell2025basicreproductionnumberpetri} developed a generalized algorithm for computing $R_0$ from any compartmental epidemic model represented as a Petri Net. Their approach derives the transmission and transition matrices symbollically from the net's incidence matrix and transition rate functions, enabling direct construction of the next generation matrix. They validate on several standard models including the SIR, SEIR, and Covid-19 models, demonstrating consistency with classical results and highlighting the advantages of compositional, structure-driven analysis.

To verify rigor and practicality, Reckell et al. \cite{reckell2025numericalcomparisonpetrinet} presented the first systematic numerical evaluation of Petri Net-based SIRS models relative to the classical ODE formulation. They compared a deterministic discrete-time Petri Net with dynamic arc weights and a stochastic, Continuous-Time Markov-chain (CTMC) Petri Net in GPenSIM and Spike tools respectively. For the deterministic model, they explored the effects of sampling frequency (i.e. 40-80 steps per ODE time unit) and rounding schemes for non-integer arc weights, introducing a novel residual-carrying method that achieves a Relative Root Mean Squared Error (RRMSE) below $1\%$ across a wide range of parameters. In the stochastic model, they demonstrated that increasing the population scale $(\geq 100)$ is sufficient to suppress stochastic variability and closely approximate the ODE solution. Their findings highlight the numerical conditions under which both Petri-net approaches can reliably mirror ODE dynamics, providing a foundation for accurate and scalable epidemiological modeling using discrete-event formalisms.

These advances in Petri Net formalisms have established a powerful, fully generative framework for both deterministic and stochastic epidemic simulations. To fully leverage this framework, accurate parameter estimation is essential. We now review key simulation-based inference methods that support parameter learning in complex epidemiological models.

\subsection{Simulation-Based Inference}
Parameter estimation is a central challenge in computational biology and epidemiology, where accurate model predictions depend on identifying underlying rates and transition probabilities from observed data. This inference problem has largely proceeded with models formulated as systems of ODEs or Stochastic Differential Equations (SDEs) using classical likelihood-based methods or Approximate Bayesian Computation (ABC). More recently, deep learning approaches have begun to bridge the gap between mechanistic and data-driven modeling. For instance, Yazdani et al.\cite{yazdani2020systems} embed ODE dynamics directly into neural network architectures allowing backpropagation through numerical solvers to recover both hidden states and rate parameters from sparse, noisy data. The work of Gaskin et al. \cite{gaskin2023neural} extended this paradigm and proposed a pipeline that combines large-scale, multi-agent ODE/SDE forward solvers with a neural network model trained on simulated trajectories. They demonstrated this on a synthetic SIR dataset, yielding full parameter-density estimates in less time while maintaining prediction quality an order of magnitude better than classical Bayesian methods. Specific to infectious disease modeling, Radev et al. \cite{radev2021outbreakflow} proposed OutbreakFlow, an amortized Bayesian framework based on invertible neural networks that learn to map noisy COVID-19 time-series directly to joint SEIR-model posteriors, enabling daily updating of epidemiological parameters with minimal overhead.

Despite these advances in simulation-based inference for ODE and agent-based approaches, parameter estimation for discrete-event Stochastic Petri Net (SPN) models remains unexplored. The unique characteristics of SPNs pose challenges for standard inference techniques that assume continuous dynamics or tractable likelihoods. Computing likelihoods in SPNs involves summing over every possible firing sequence which grows exponentially with the number of transitions. This ends up with a combinatorially exploding likelihood with no closed form \cite{ciardo1995discrete}. This motivates the need for novel, likelihood-free data-driven surrogate methods capable of estimating parameters directly from observed trajectories, without relying on explicit likelihood functions. Having identified this gap, we now formalize the problem setting and present our proposed neural-surrogate methodology.

\begin{figure}
    \centering
    \includegraphics[width=0.95\linewidth]{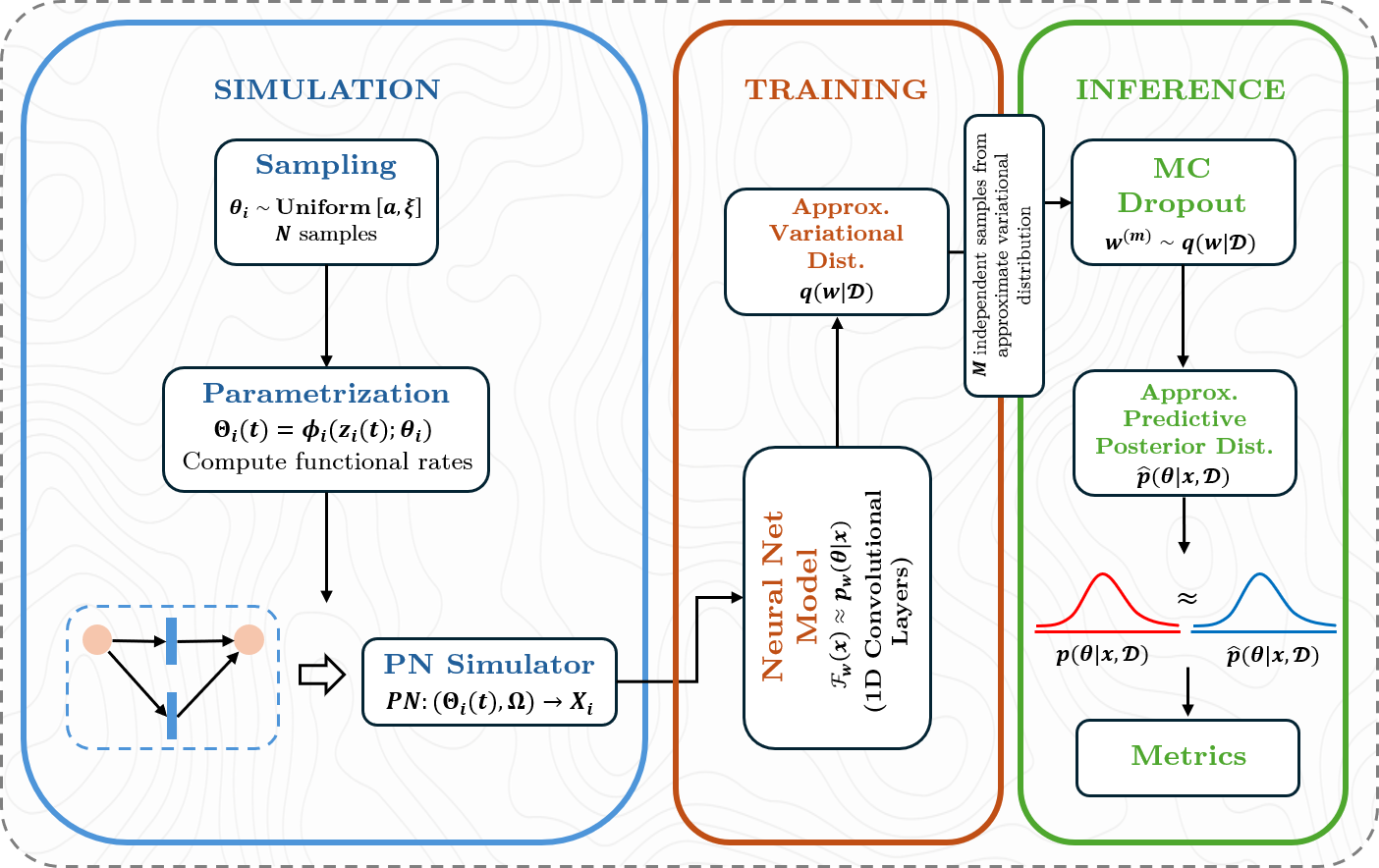}
    \caption{Method Pipeline}
    \label{fig:enter-label}
\end{figure}

\section{Method}
\subsection{Problem Setting}
Let $\mathcal{D}=\{(X_i,\theta_i)\}_{i=1}^N$ be an $iid$ dataset of $N$ simulated epidemic scenarios, where each $X_i\in\mathbb{R}^{T\times d_{in}}$ represents a single time-indexed simulation output over $T$ timesteps and $\theta_i\in\mathbb{R}^k$ is a vector of underlying coefficients used to define the model's functional parameters for that sample.

Suppose $\mathcal{PN}$ represents a Vector-Patch Petri Net model, $\Theta_i$ represents the functional model transition rates with underlying coefficients $\theta_i$, $y_i$ represents the environmental covariates and $\phi_i$ represents the basis functions that model the covariates (discussed in Section 2). Then each $X_i$ is the output of a Vector-Patch Petri Net based epidemic model $\mathcal{PN}$, which encodes stochastic disease dynamics using discrete places and transitions. The model evolves over time governed by time-varying functional parameters, defined by the interaction of covariates $y_i$ with the coefficient vector $\theta_i$, given by $$\Theta_i(t)=\phi_i(y_i(t);\theta_i)$$ In addition, the PN simulator also depends on a fixed vector of non-varying parameters $\Omega\in\mathbb{R}^q$ across the $N$ samples. Thus, the PN simulator produces a stochastic time series which can be modeled as $$X_i = \mathcal{PN}(\Theta_i(t),\Omega)+\varepsilon_i, \quad\quad \varepsilon_i\sim \mathcal{N}(0,\sigma^2\mathbb{I})$$ where  $\varepsilon_i$ reflects Gaussian stochastic fluctuations, for example due to observation errors or unobserved processes.

Our objective is to solve an inverse problem and learn a surrogate model that, given a new observed trajectory $x$, infers the underlying parameter vector $\theta$, and quantify the uncertainty in the estimates. To this end, we consider a neural network model $\mathcal{F}_w(x)$ that parameterizes a predictive distribution $$\mathcal{F}_w(x)\approx p_w(\theta|x)$$ where $w$ are the learnable weights. In a Bayesian setting, this corresponds to approximating the posterior predictive distribution
\begin{equation}
    p(\theta|x,\mathcal{D})=\int p(\theta|x,w)~p(w|\mathcal{D})~dw
\end{equation}
where $p(w|\mathcal{D})$ denotes the posterior distribution over network weights.

We describe in Section \textit{D} how this predictive distribution is efficiently approximated using Monte Carlo dropout \cite{gal2016dropout}, enabling uncertainty-aware inference with scalable training.

\subsection{Forward Model (Petri Net Model)}
The forward model is a mapping from known functional and fixed parameters $(\Theta_i(t),\Omega)$ to observations $X_i$ across time. The model is a two-patch host-vector epidemic discrete time Stochastic Petri Net. Each patch $i$ and $j$ have the following places; Susceptible Humans $(S_H)$, Infected Humans $(I_H)$, Recovered Humans $(R_H)$, Susceptible Mosquitoes $(S_M)$, Infected Mosquitoes $(I_M)$ and their respective dead places. For each patch $i$ or $j$, we set initial tokens or populations as $S_H=4000$, $I_H=20$, $R_H=20$, $S_M=2000$ and $I_M=10$. All the initial respective dead places were set to $0$. The parameters of the model for each patch and between patch include Mosquito-to-Human transmission rate, Human-to-Mosquito transmission rate, Human mortality rate, Mosquito mortality rate, Human migration rate, Mosquito migration rate, Human recovery rate denoted by $\beta_{HM},~\beta_{MH},~\mu_H,~\mu_M,~\varphi,~\alpha,~\sigma$ respectively. These parameters are embedded in the transitions between the places of the Petri Net model which represent the arc weights of the model.

The initial parameter values are displayed in Table \ref{init_model_params}. (Notice that subscript $ii$ represents a same-patch parameter and subscript $ij$ represents a between-patch parameter.) All initial parameter values are the same for each patch. Among these parameters, $\beta_{MH},~\beta_{HM},~\mu_M$ and $\alpha$ are considered as functional parameters with constants $\theta_i$ and are discussed further in the next sections.

\begin{figure}
    \centering
    \includegraphics[width=0.9\linewidth]{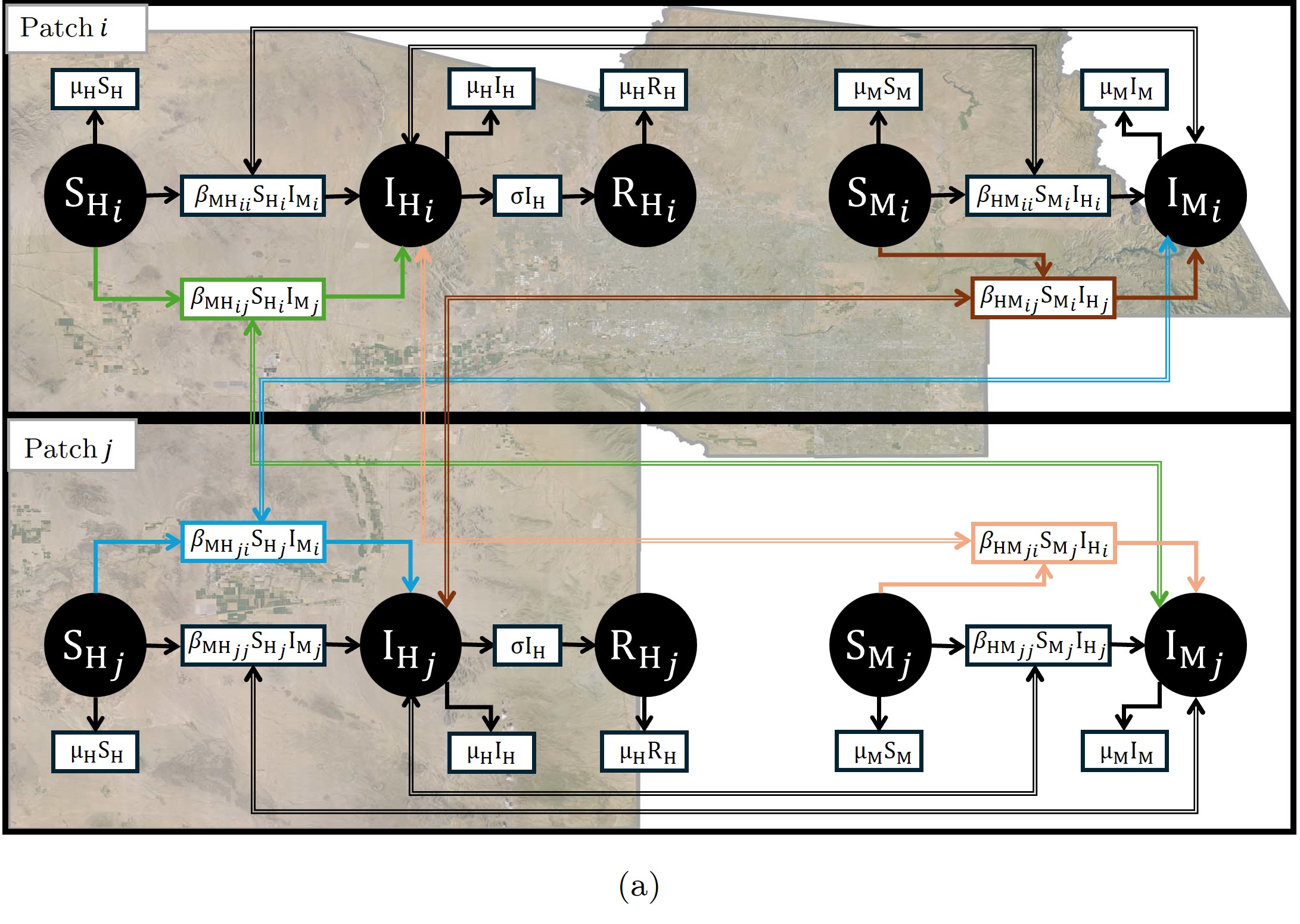}
    \label{fig:enter-label}
\end{figure}
\begin{figure}
    \centering
    \includegraphics[width=0.9\linewidth]{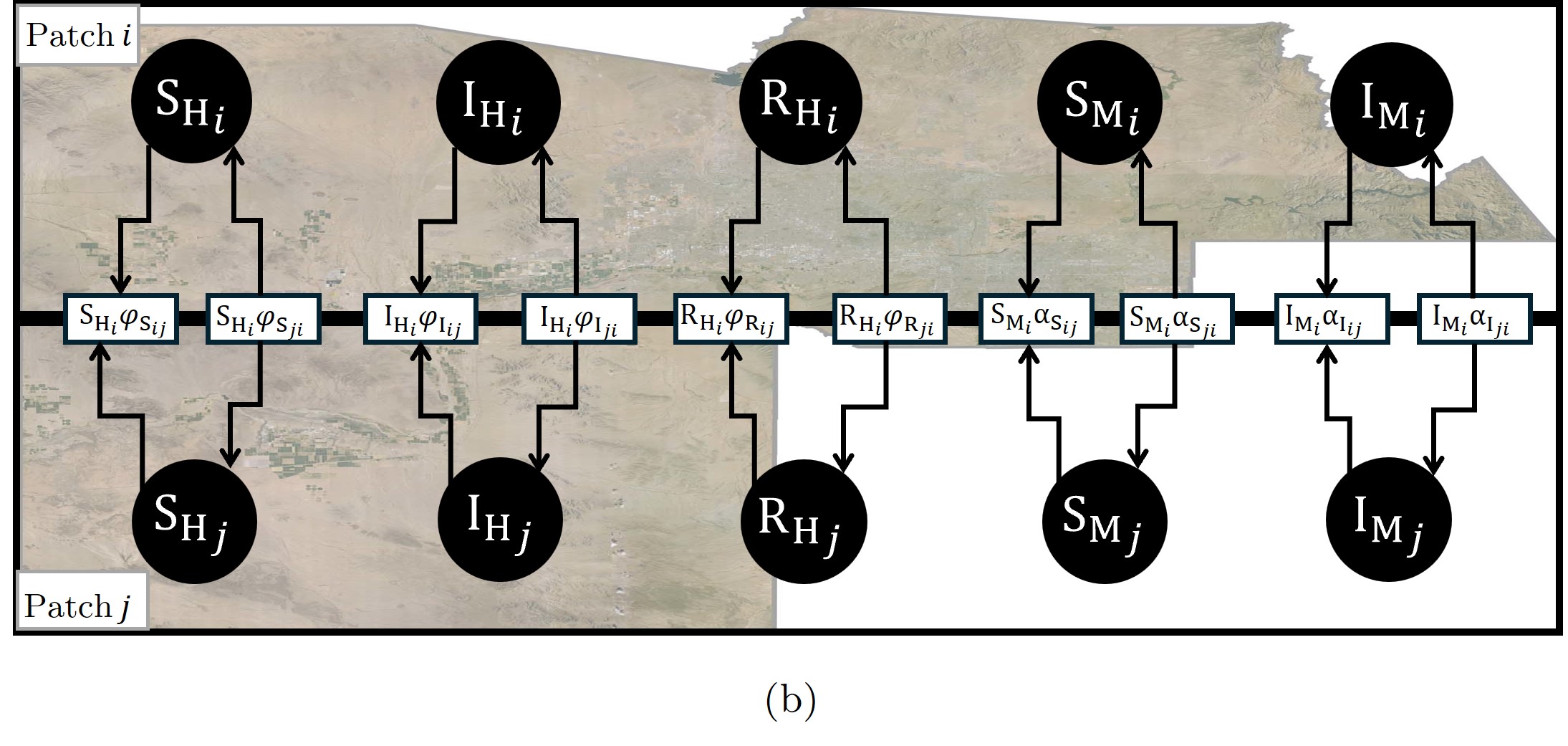}
    \caption{(a) Represents within and between patch infection, recovery and mortality dynamics (b) Represents between patch migration dynamics (Note: (a) and (b) together form one model). Double arcs indicate tokens to and from a place and transition}
    \label{fig:enter-label}
\end{figure}

\begin{table}[width=.9\linewidth,cols=4,pos=h]
\caption{Initial Model Parameters}
\label{init_model_params}
\begin{tabular*}{\tblwidth}{@{} LLLL@{} }
\toprule
\textbf{Parameter} & \textbf{Value}\\
\midrule
$\beta_{MH_{ii}}$ & $0.30$\\
$\beta_{HM_{ii}}$ & $0.40$\\
$\beta_{MH_{ij}}$ & $0.20$\\
$\beta_{HM_{ij}}$ & $0.25$\\
$\sigma$ & $0.10$\\
$\mu_H$ & $0.01$ \\
$\mu_M$ & $0.02$ \\
$\varphi_S$ & $3.51e-4$\\
$\varphi_I$ & $3.50e-4$\\
$\varphi_R$ & $3.49e-4$\\
$\alpha_S$ & $3.40e-4$\\
$\alpha_I$ & $3.40e-4$\\
\bottomrule
\end{tabular*}
\end{table}

It is worth noting that our main goal is to show the neural surrogate can recover the functional dependence of the latent constants or coefficients on the exogenous environmental variables, not to investigate the accuracy of the simulation model we use for any particular system. In other words, our focus is on inferring the underlying coefficients as an essential methodological step to fitting and comparing Petri Net models of infectious disease systems whose parameters exhibit spatial-temporal dependence on environmental variables.

\subsubsection{Information Layers Integration}
In a standard Petri Net formalism, system evolution is determined by the firing of transitions, each governed by its arc weights which represents the number of tokens consumed or produced during state changes. We extend the traditional Petri Net formalism to account for environmental impacts on mosquito-borne disease transmission dynamics by integrating environmental covariate layers. These layers reflect time-varying temperature and relative humidity, which influence the frequency or severity of disease infections, transmissions, mortality, and migration. Rather than altering the arc weights directly, we indirectly alter the rates (for example, we alter $\beta_{HM}$ in $\beta_{HM} SI$) as a function of daily changes in environmental conditions. This is discussed in detail in the Section 2.

To illustrate how the neural surrogate can work on realistic environmental data, we obtained 2022 Maricopa County daily temperature and relative humidity data segregated by all the weather stations in the county. Temperature values were originally reported in degrees Fahrenheit and were converted to degrees Celsius using the formulation $T_{Celcius}=\frac{5}{9}(T_{Fahrenheit}-32)$. The data was linearly interpolated in time to estimate missing daily values as well as smoothen over any temporal gaps in the data. To reduce spurious day-to-day noise in the data while keeping underlying seasonal patterns, we applied a centered $7$- day moving average to both temperature and relative humidity extremes before proceeding with any further processing.

To derive daily patch-specific environmental data, we overlaid the geographic locations of weather stations on the Maricopa county polygon and divided the study area into two spatial patches. Weather stations located within each patch were identified and for each day, the average of all available station-specific values within a given patch was computed. This average was used to represent the daily environmental condition of that patch in the model. This approach provides a spatially aggregated yet locally representative estimate of environmental exposure for each modeled region. While we use only two coarse-grained patches in our study, the general approach could be extended to include many patches representing smaller areas.

\subsubsection{Mathematical Formulations}
 For each patch and day $d$, let $T_d$ represent the observed daily average temperature, and $RH_d^{min}, ~ RH_d^{max}$ the daily minimum and maximum relative humidity respectively. We therefore employ the following mathematical formulations as basis functions for our use case.
 \begin{enumerate}
     \item[i)] \textit{Transmission Thermal Response (Briere)} \cite{briere1999novel}: We model the temperature-dependent biting rate $B(T_d)$ of mosquitoes with the Briere function. Over the biologically plausible temperature range $\mathcal{T}=[T_{min},T_{max}]$, we define the function as
     \begin{equation}
        B(T_d) =\begin{cases} aT_d(T_d-T_{min})(T_{max}-T_d)^{\frac{1}{2}}, \\ \quad\quad\quad\quad T_{min}<T_d<T_{max}\\ 0, \quad \text{otherwise} \end{cases}
     \end{equation}
     where $a>0$ and $T_{min}<T_{max}$ are set within biologically viable thermal bounds for biting activity of mosquitoes.
     
     To avoid zero-flooring and account for diurnal variation of the Briere development curve when $T_d<T_{min}$ or $T_d>T_{max}$, we adopt the diurnal integration technique described in \cite{mordecai2017detecting} and proposed by \cite{parton1981model}. For day $d$ with observed daily mean $T_d$, minimum $T_d^{min}$ and maximum $T_d^{max}$, we reconstruct an hourly temperature profile given by $$T_{d,h}=T_d+\frac{T_d^{max}-T_d^{min}}{2}\sin\Big(\frac{2\pi}{24}h-\frac{\pi}{2}\Big)\quad (*)$$ where $0\leq h\leq 23$. We then compute the Briere response at each hour and average over 24 hours.
     
     \item[ii)] \textit{Temperature-Dependent Mortality:} We describe the temperature effects on mosquito mortality by adapting the unimodal quadratic function for mosquito egg-to-larva survival proposed by \cite{mordecai2017detecting}. This function assumes that there is high mortality below a critical lower threshold $T_{0}$ and above an upper threshold $T_{m}$ where lowest mortality occurs at an intermediate optimum. In its general form, we have
     \begin{equation}
         Q(T_d) = -c(T_d-T_{0})(T_d-T_{m})
     \end{equation}
     where $T_d$ is the daily temperature, $T_0$ is the minimum temperature for survival, $T_m$ is the maximum temperature for survival and $c$ is a scaling parameter. Parameters $T_0$, $T_m$ and $c$ were used as estimated in their work and are displayed in Table II.
     \item[iii)] \textit{Humidity-Driven Response:} To model the effects of relative humidity on mosquito activity, we incorporate a logistic scaling function as follows
     \begin{equation}
         L(RH_d)=\frac{1}{1+\exp[-k(RH_d-RH_{opt})]}
     \end{equation}
     where $k>0$ is the humidity sensitivity and $RH_d$ is the daily average relative humidity $(\%)$ in a patch. This form reflects increased movement and biting propensity under moist conditions and align with empirical observations of desiccation-limited vector behavior \cite{brown2023humidity}.
     
     Prior to this, we also reconstruct a continuous hourly relative humidity profile from the daily minima and maxima. As such, we model the hourly humidity $RH_{d,h}$ by a half cosine curve as follows. $$RH_{d,h}=RH_d^{min}+\frac{RH_d^{max}-RH_d^{min}}{2}\Big[1-\cos\Big(2\pi\frac{h}{24}\Big)\Big]$$ We then sample the reconstructed relative humidity every four hours, apply the logistic transform using equation (4) and average the resulting six values. This guarantees that the nonlinear humidity impact is integrated across the whole diurnal cycle, capturing peaks and troughs in $RH$.

     \item[iv)] \textit{Diffusion Coefficient:} To capture the temperature-relative humidity dependence of migration or flight activity, we considered a piecewise function with temperature thresholds and a relative humidity-driven response within the thermal window.
     \begin{equation}
         D(T_d,RH_d)=\begin{cases}
             L(RH_d), & \quad T_{-}<T_d<T_{+}\\
             0, & \quad \text{otherwise}
         \end{cases}
     \end{equation}
     where $T_{-}$ and $T_{+}$ are the minimum and maximum temperature thresholds for mosquito flight activity. $L(RH_d)$ is the relative humidity response as seen in Equation 4. We adopted fitted values for $T_{-}$ and $T_{+}$ from \cite{reinhold2018effects} and are shown in Table II.\\

     In addition, we also define a kernel migration function to take into account the distance between patches based on the max flight distance of mosquitoes. Thus, to model mosquito migration across the true minimal gap between patches, we do simple computation of pair-patch distance $\Delta_{ij}$ using the Haversine distance. This approach finds the distance between the closest points of two adjacent or non-adjacent patches. If two patches are adjacent, then $\Delta_{ij}=0$ trivially. We set a flight distance cut-off to 3 since the max flight of most mosquito species is between 1-3 miles \cite{verdonschot2014flight}. Thus, we can define the mosquito migration kernel function as follows
    $$d_{M_{ij}}^{(c_{\{S,I\}})}=\begin{cases}\frac{1}{c_{\{S,I\}}+\Delta_{ij}^2},\quad \Delta_{ij}\leq 3\\0, \quad\quad \Delta_{ij}>3 \end{cases}$$ where $S$ and $I$ are the susceptible and infected mosquito population migrating between patches respectively. $c_{\{S,I\}}$ unfolds to $c_S$ and $c_I$ which are just constants in the kernel function.
    This assumes that mosquito population is distributed homogeneously within a patch. Since our vector-patch Petri Net model has only two patches, $\Delta_{ij}=0$ leaving the migration rate dependent only on the constant $c$. However, this formulation is especially valuable in multi-patch Petri Net models (i.e. with $\geq 2$ patches).
 \end{enumerate}

\begin{table*}[cols=3,pos=h]
\caption{Calibrated Parameters ($k$ is fixed)\\ (All parameters represent the \textit{Aedes species})}
\label{calib_params_lit}
\begin{tabular*}{\tblwidth}{@{} LLL@{} }
\toprule
\textbf{Parameter} & \textbf{Description} & \textbf{Value}\\
\midrule
$a$~\cite{mordecai2017detecting} & Biting scale parameter & $2.71e-04$\\
$T_{min}$~\cite{mordecai2017detecting} & Minimum temperature (Biting) & $14.67^oC$ \\ 
$T_{max}$~\cite{mordecai2017detecting} & Maximum temperature (Biting) & $41^oC$\\ 
$k$ & Humidity sensitivity & $0.1$ \\
$RH_{opt}$~\cite{brown2023humidity} & Optimal relative humidity & $70\%$ \\
$c$~\cite{mordecai2017detecting} & Survival scale parameter & $-3.36e-03$  \\
$T_0$~\cite{mordecai2017detecting} & Minimum Temperature (Survival) & $7.68^oC$ \\
$T_m$~\cite{mordecai2017detecting} & Maximum Temperature (Survival) & $38.31^oC$ \\
$T_{-}$~\cite{reinhold2018effects} & Minimum Temperature (Diffusion) & $10^oC$ \\
$T_{+}$~\cite{reinhold2018effects} & Maximum Temperature (Diffusion) & $35^oC$ \\
\bottomrule
\end{tabular*}
\end{table*}

Each of these temperature-humidity dependent basis functions' internal parameters are either drawn from literature or fixed (See Table \ref{calib_params_lit}). These values are not intended to represent any one real specific \textit{Aedes} mosquito species or genus, but rather to enable biologically plausible simulations.

\subsubsection{Parametrization and Sampling}
We grounded our transmission and mortality rates in the entomological ranges reported in \cite{jin2020mathematical}. Based on their work, our \textit{Mosquito-to-Human transmission rate} $\beta_{MH}\in[0.01, 0.80]~~\text{day}^{-1}$, \textit{Human-to-Mosquito transmission rate} $\beta_{HM}\in[0.072, 0.64]~~\text{day}^{-1}$ and \textit{Mosquito mortality rate} $\mu_M\in[0.05, 0.33]~~\text{day}^{-1}$. To capture environmental modulation by temperature and relative humidity, we expressed each rate as a linear combination of the basis functions $B(T_d), Q(T_d)$ and $L(RH_d)$ discussed in Section 2, which by construction, each takes values in $[0,1]$.
 
 Let $i,j\in\{1,2\}$ represent patch indices. Then for each day $d$ and patch $i,j$, we compute the PN model functional parameters as follows.
 \begin{enumerate}
     \item[i)] \textit{Within-Patch Transmission Rate} (i.e. if $i=j$): We compute the Mosquito-to-Human and Human-to-Mosquito transmission rate within a patch as follows.\\
     \begin{equation}
        \begin{split}
        \beta_{MH_{ii}}(T_{d_i},RH_{d_i}) &= \lambda_0 + \lambda_1 B_b(T_{d_i}) + \lambda_2 L(RH_{d_i})
        \end{split}
     \end{equation}
     \begin{equation}
        \begin{split}
            \beta_{HM_{ii}}(T_{d_i},RH_{d_i}) & = \gamma_0 + \gamma_1 B_b(T_{d_i}) + \gamma_2 L(RH_{d_i})
        \end{split}
     \end{equation}
     \item[ii)] \textit{Between-Patch Transmission Rate} (i.e. if $i\neq j$): We also compute the transmission rate between the two patches in cases where infectious mosquitoes come into contact with humans and vice versa. In our computations, we use the destination patch covariate conditions i.e. the patch where the biting happens after contact. For example, $\beta_{MH_{ij}}$ means transmission rate as a result of an infectious mosquito from patch $j$ interacting (eg. biting) with a susceptible human in patch $i$. The equations are as follows.
     \begin{equation}
        \begin{split}
         \beta_{MH_{ij}}(T_{d_j},RH_{d_j}) & = \delta_0 + \delta_1 B_b(T_{d_j}) + \delta_2 L(RH_{d_j})
        \end{split}
     \end{equation}
     \begin{equation}
        \begin{split}
            \beta_{HM_{ij}}(T_{d_j},RH_{d_j}) & = \eta_0 +\eta_1 B_b(T_{d_j}) + \eta_2 L(RH_{d_j})
        \end{split}
     \end{equation}
     \item[iii)] \textit{Mosquito Mortality Rate}: We also compute the Mosquito mortality rate for each patch as
     \begin{equation}
         \mu_{M_i}(T_{d_i},RH_{d_{i}}) = \rho_0 + \rho_1 Q(T_{d_{i}}) + \rho_2 L(RH_{d_{i}})
     \end{equation}
\end{enumerate}

 \begin{enumerate}
     \item[iv)] \textit{Mosquito Migration Rate:} To account for spatial dynamics in the mosquito population, we define the migration rate $\alpha_{M_{ij}}$ as the rate of movement of mosquitoes from patch $i$ to patch $j$. Migration is assumed to be asymmetric (i.e. $i\neq j$) and is dependent on both environmental conditions in the source patch and the distance between patches. In addition, we distinguish between the movement patterns of susceptible and infected mosquitoes.
     Thus, the migration rates of susceptible and infected mosquitoes between patches can be defined as
     \begin{equation}
         \alpha_{M_{ij}}^{(S)}(T_{d_i},RH_{d_i}) = d_{M_{ij}}^{(c_S)}\cdot \iota_1 D(T_{d_i},RH_{d_i})
     \end{equation}
     \begin{equation}
         \alpha_{M_{ij}}^{(I)}(T_{d_i},RH_{d_i}) = d_{M_{ij}}^{(c_I)}\cdot \iota_2 D(T_{d_i},RH_{d_i})
     \end{equation}
 \end{enumerate}

 The internal coefficients $\lambda_0,~\gamma_0,~\delta_0,~\eta_0,~\rho_0$ are baseline rates under reference conditions. $\lambda_k,~ \gamma_k,~ \delta_k,~ \eta_k, ~\rho_k$ for all $k\in\{1,2\}$ are the single-term coefficients that weights how strongly each covariate modulates transmission or mortality.
 
 For our sampling procedure for the coefficients $\lambda_k$, $\gamma_k$, $\delta_k$, $\eta_k$, $\rho_k$, $\iota_k$ where $k\in\{0,1,2,3\}$, we ensure that for any $B(T_d), Q(T_d), L(RH_d)\in[0,1]$, the rates remain within their entomological bounds by sampling these coefficients as follows. If $k=0$, $\lambda_k,~ \gamma_k,~ \delta_k,~ \eta_k$, and $\rho_0$ are chosen as the minimum bounds of respective rates.
 
 For $\beta_{MH}, ~\beta_{HM}$, and $\mu_M$, we compute the total gain $\xi_{MH},~\xi_{HM}$, and $\xi_{\mu}$ respectively as the difference of the maximum and minimum bounds. If $k=1,2,3$, we then sample the remaining coefficients $\lambda_k,~ \gamma_k,~ \delta_k,~ \eta_k$, and $\rho_1$ from a uniform distribution on $[0,\xi]$ where $\xi$ is the corresponding rate's total gain. The migration rates internal coefficients $\iota_k$ were also sampled from a uniform distribution on $[0.06,0.6]$ based on the work of \cite{juarez2020dispersal, ciota2012dispersal, costantini1996density}. This construction guarantees the rates to fall in the specified ranges.

 \subsection{The Inverse Model (Neural Network)}
We propose a 1D convolutional residual network (1D-ResNet) for parameter estimation. Let each input sample be $x\in\mathbb{R}^{T\times d_{in}}$. Our model first permutes each input to channels-first, $x_{perm}\in\mathbb{R}^{d_{in}\times T}$, helping to reduce memory overhead during computations. We then project via a $1\times 1$ convolution to $F$ filters, learning an optimal linear combination of original feature channels at each timestep before deeper processing, followed by a Rectified Linear Unit (ReLU) activation: $$x^0=\text{ReLU}(W^{(0)}*x_{perm}+b^{(0)})$$ where $*$ denotes a 1D convolution. We then stack $B$ residual blocks such that for $i=1,...,B$, we have $$r^i=\text{dropout}(\text{ReLU}(W^{(i)}*x^{i-1}+b^{(i)})), \quad x^i=x^{i-1}+r^i$$ Each convolutional layer has kernel size $k$, and a dropout rate $p$, both in the hidden convolutional layers and before the final head. The residual block \cite{he2016deep} is important for preventing the problem of vanishing gradients by adding the skip connection $x^i$. It ensures stable gradient flow, enabling our model to learn incremental feature refinements rather than full re-mappings in each block.

We then apply a global average pooling (GAP) over time, collapsing the temporal dimension into one vector, and apply a final ReLU, dropout, and fully-connected layer to map to the $d_{out}$ outputs. To ensure our outputs $\theta\in[0,1]$, we apply a sigmoid activation at the final stage. Mathematically, $$z=\text{dropout}(\text{ReLU}(\text{GAP}(x^B))), \quad \theta=\text{Sig}(zW^{fc}+b^{fc})$$ For our model, we set $F=128$ filters, a kernel size $k=5$, $B=3$ residual blocks, $p=0.1$ dropout and $d_{out}=12$ (since we are estimating 12 parameters).

There are two reasons why we include dropout layers, first to regularize the network and reduce overfitting during training, and most importantly for the purpose of this work to quantify epistemic or model uncertainty by keeping it active at inference time, which we discuss in the next section.

\subsection{Monte Carlo Dropout (MC Dropout)}
Dropout in neural networks can be interpreted as performing approximate variational inference in a Bayesian deep Gaussian process, as shown in \cite{gal2016dropout}. Directly evaluating equation (1) is intractable for deep nets, so we follow a simple approach proposed by \cite{gal2016dropout}.

Specifically, each stochastic forward pass with dropout at inference time corresponds to drawing $M$ independent Monte Carlo samples from an approximate variational posterior distribution $q(w|\mathcal{D})$ over the model weights (i.e. $w^{(m)}\sim q(w|\mathcal{D})$). We then compute $\theta^{(m)}=\mathcal{F}_{w^{(m)}}(x)$ for each pass. This allows us to approximate the predictive distribution over the parameters $\theta$ furnishing us with a Monte Carlo approximation to the posterior predictive distribution $$\hat{p}(\theta|x,\mathcal{D})\approx \frac{1}{M}\sum_{m=1}^M p(\theta|x,w^{(m)})$$

We estimate the predictive mean as
\begin{equation}
    \mathbb{E}_{\hat{p}(\theta|x,\mathcal{D})}(\theta)\approx \frac{1}{M}\sum_{m=1}^M \mathcal{F}_{w^{(m)}}(x)
\end{equation}
and the predictive variance as
\begin{equation}
    Var_{\hat{p}(\theta|x,\mathcal{D})}(\theta)\approx \frac{1}{M-1}\sum_{m=1}^M\Big(\mathcal{F}_{w^{(m)}}(x)- A\Big)^2
\end{equation}
where $A=\displaystyle\frac{1}{M}\sum_{m=1}^M \mathcal{F}_{w^{(m)}}(x)$\\ 
In practice, the predictive variance is used to estimate the epistemic uncertainty (model uncertainty) while the predictive mean represents the center of our predictive distribution. For our use case, we set $M=50$ stochastic forward passes with dropout $p=0.1$ kept at inference time for each test example.

\subsection{Metrics}
To assess both point-estimate accuracy and uncertainty quantification in our NN MC-dropout parameter recovery model, we report the following metrics for each parameter $\theta_i$ computed over $N_{test}$ test samples. First we compute the average of the true parameter values over all test samples (denote by $\bar{\theta}_i$) and the average of the model's posterior mean predictions using equation (13) (denote by $\bar{\hat{\theta}}_i$). We compute $|\text{Bias}|=|\bar{\hat{\theta}}_i-\bar{\theta}_i|$, a measure of systematic error i.e. how far on average our predicted parameters are from the true parameters. We use the Root Mean Squared Error (RMSE) to quantify the overall prediction error given by $$\text{RMSE}_i=\sqrt{\frac{1}{N_{test}}\sum_{j=1}^{N_{test}}(\hat{\theta_i}-\theta_i)^2_j}$$

For uncertainty estimates, we employ equation (14) and compute its average over all test samples given by 
\begin{equation}
    \bar{\hat{\sigma}}_i=\frac{1}{N_{test}}\sum_{j=1}^{N_{test}} \sqrt{Var_{\hat{p}(\theta_i|x,\mathcal{D})}(\theta_i)}_{j}
\end{equation} where $Var_{\hat{p}(\theta_i|x,\mathcal{D})}(\theta_i)_{j}$ is the MC-dropout posterior variance for sample $j$.

Note that for better interpretability, we use the standard deviation instead of variance, that is, we take the square root of equation (14) for each sample as seen in equation (15)). This gives us the model's uncertainty in predicting each parameter $\theta_i$.

We further assessed the fraction of times the true parameter falls within the model's $\pm 1\sigma$ interval by computing the per-parameter Coverage as 
\begin{equation}
    \frac{1}{N_{test}}\sum_{j=1}^{N_{test}}\textbf{1}\Big(|\hat{\theta}_i-\theta_i|_j\leq \sqrt{Var_{\hat{p}(\theta_i|x,\mathcal{D})}(\theta_i)}_{j}\Big)
\end{equation}
For a well-calibrated Gaussian uncertainty, we expect $\approx 68\%$ coverage. Despite our model achieving a very low prediction error, we observed that coverage was substantially below or above this threshold indicating high over-confidence or under-confidence in predictions respectively. Thus, uncertainty estimates were not initially well-calibrated.

\subsection{Predictive Uncertainty Recalibration}
We further assessed and re-calibrated our predictions following the STD-scaling method proposed by \cite{levi2022evaluating} to ensure that our model's uncertainty estimates accurately reflect its empirical errors. Their approach multiplies all predicted standard deviations by a scalar $s$. In our case, we have $s_i$ $\bar{\hat{\sigma}}^{\text{cal}}_i = s_i \cdot \bar{\hat{\sigma}}$ per parameter. This scalar $s_i$ is chosen based on which parameter the model is over or under-confident in predicting. We set $s_i>1$ for a parameter the model is over-confident in predicting and set $s_i<1$ otherwise. The scaling factor $s_i$ was estimated on our held-out validation set by minimizing the Gaussian Negative log-likelihood (NLL) given by
\begin{equation}
    \hat{s}_i = \underset{s_i}{\text{argmin}}\bigg(\frac{N_{val}}{2}\log(s_i)-\sum_{j=1}^{N_{val}}\frac{( \bar{\hat{\theta}}_i-\bar{\theta}_i)^2}{2(s_i\bar{\hat{\sigma_i}})^2}\bigg)
\end{equation}
Once this scaling factor is learned, it is then applied to all predicted standard deviations on the test set. Most importantly, this recalibration does not affect the predicted means ensuring only the rescaling of the predicted standard deviations.

For our post-recalibration evaluation, we plotted both the pre- and post-recalibration nominal versus empirical coverages for multiple confidence levels i.e. $50\%, ~68\%, ~90\%, ~95\%$ for each parameter. We further evaluated the calibration quality by first grouping all test predictions into $K=10$ bins according to their predicted standard deviations. For each bin $\mathcal{B}_k$, we computed
$$(\text{RMV}_k)_i=\sqrt{\frac{1}{|\mathcal{B}_k|}\sum_{j\in\mathcal{B}_k}(\bar{\hat{\sigma}}_i)_j}$$ $$(\text{RMSE}_k)_i=\sqrt{\frac{1}{|\mathcal{B}_k|}\sum_{j\in\mathcal{B}_k}(\hat{\theta_i}-\theta_i)^2_j}$$
the Root Mean Variance (RMV) (the average predicted uncertainty within the bin and RMSE (the empirical error of the model within the same bin) following \cite{levi2022evaluating}. For a perfectly calibrated model, we expect the $(\text{RMSE}_k)_i\approx (\text{RMV}_k)_i$. This furnished us with a reliability diagnoses of our model. To summarize this deviation or error in calibration, we computed the Expected Normalized Calibration Error (ENCE) \cite{levi2022evaluating} for each parameter as follows $$\text{ENCE}_i=\frac{1}{K}\sum_{k=1}^K\frac{|(\text{RMSE}_k)_i-(\text{RMV}_k)_i|}{(\text{RMV}_k)_i}$$ This measures the average relative gap between predicted and observed uncertainty across bins.

\section{Experiment}
All simulated trajectories from our 2-Patch Vector Petri Net model were generated with the Spike Command-line tool (Spike v1.60rc2) developed by \cite{chodak2021spike}, wrapped in python 3.10. We perform 5000 independent Gillespie stepwise simulations using the direct solver with 2 runs each. Each independent sample simulation has its corresponding set of coefficients sampled uniformly as discussed in Section III. For each setting or sample, we generated 365 timesteps indicating 365 days of disease dynamics, where each day is updated with its corresponding covariate-influenced parameter. For faster results, simulations were run in parallel on 30 core CPUs, each CPU core handling a simulation run at a time, with a 15 GB memory usage on the ASU supercomputer \cite{jennewein2023sol}.

For our training setup, we first split the simulated data into 80\% train, 10\% validation and 10\% test sets. We first augment our training data with a Gaussian input noise during training with standard deviation of 0.05. This added noise is purely synthetic and not the same as the inherent simulation stochasticity $\varepsilon_i$ defined earlier in Section II (\textit{Problem Setting)}. It regularizes the network making it robust to small input-measurement errors or rounding and also helps it to generalize to unseen parameter variations.

We then train our NN model with a batch size of $8$ and run for up to $50$ epochs, optimizing with Adam with a learning rate of $1e-4$. We apply early stopping after 10 epochs without improvement. For all our experiments, we fix all random seeds to 42 for reproducibility. Our model experiment was also conducted on the ASU supercomputer with 12 CPU cores and 32 GB of memory.

Concerning computational time, each stepwise simulation sample took $\approx 196$ seconds wall time for each parallel run (each parallel simulation runs 30 samples), making a total of $196s\times 166\approx 32,732$ seconds $(\approx 9~\text{hrs}~5~\text{mins}~30s)$ for the total 5000 samples. We achieved full model training in $26~\text{mins}~6s$, and only required $54.7$ seconds for our 50 stochastic forward passes per sample during inference on our hardware, demonstrating high computational efficiency.

\section{Results and Discussion}
\begin{figure}
    \centering
    \includegraphics[width=0.95\columnwidth]{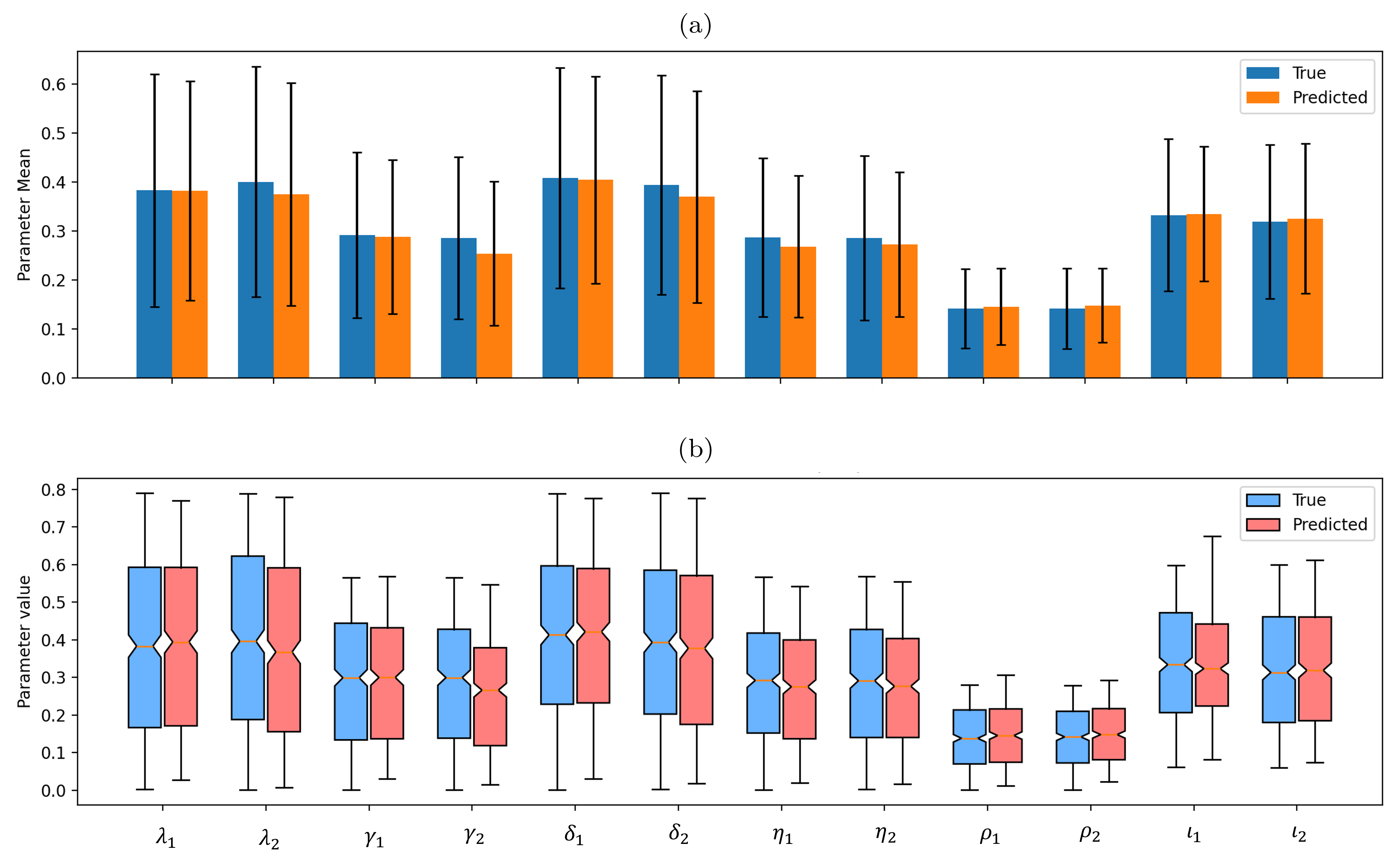}
    \caption{(a) True parameter mean compared to overall predicted parameter mean with $\pm 1\sigma$ (b) Distribution of true values across test samples compared to distribution of predicted values across test samples}
    \label{fig:bar_comparisons_param_dist}
\end{figure}

\begin{figure}
    \centering
    \includegraphics[width=0.9\columnwidth]{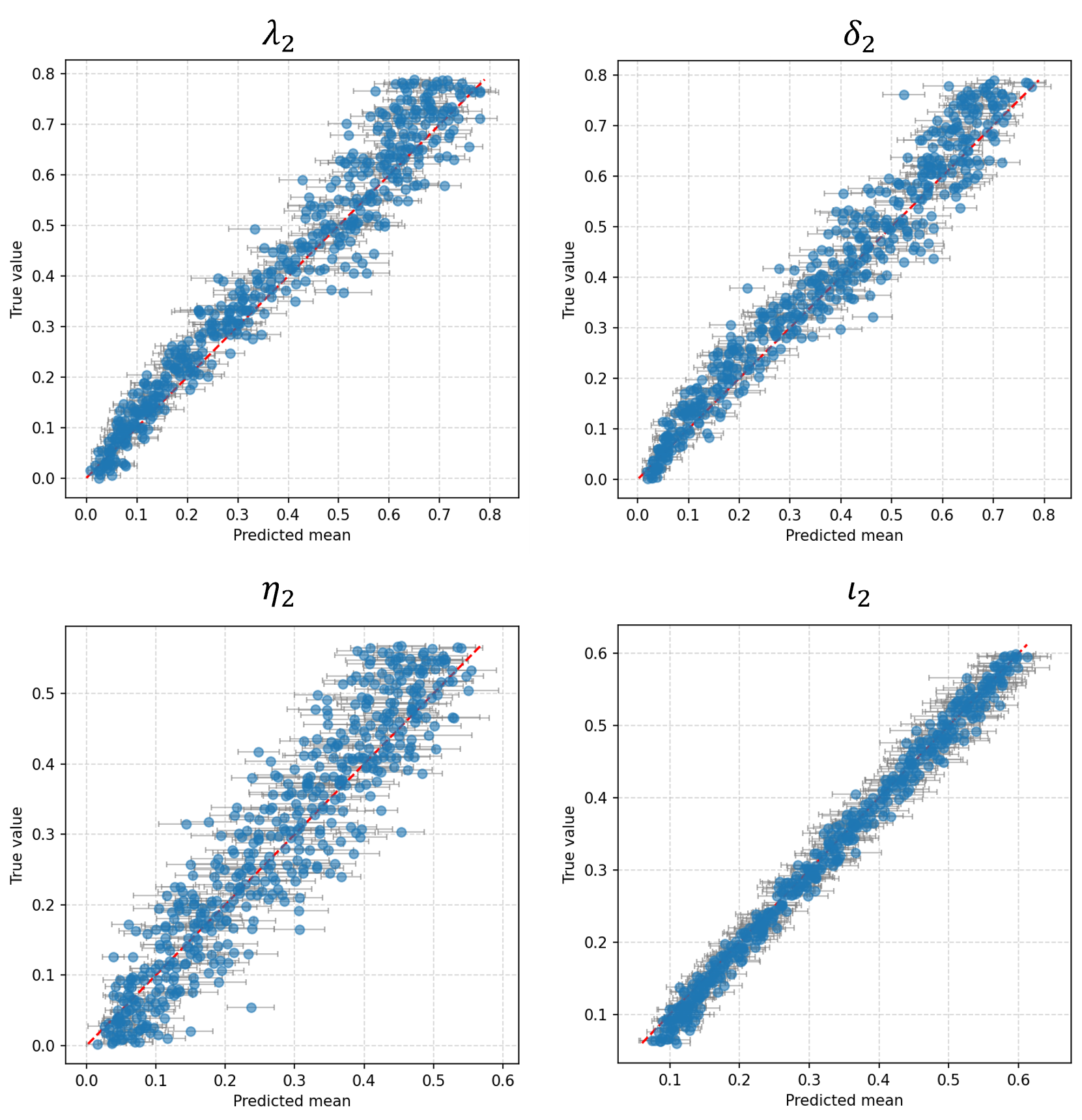}
    \caption{True and predicted sample parameters. The blue points represent the predicted sample parameters with $\pm 1\sigma$ error bars}
    \label{fig:true_vs_pred_scatter}
\end{figure}

Table \ref{results_0.1} reports per-parameter true and predicted means and bias, RMSE, average predicted standard deviation and empirical $\pm 1\sigma$ coverage on the held-out test set for both pre- and post-recalibration. Mean RMSE across all 12 parameters remains as low as 0.043, with the lowest RMSE on $\rho_1$ $(0.014)$ and the highest on $\gamma_2$ $(0.067)$. Our model's epistemic uncertainty (Avg. STD) varied from $0.018$ for $\rho_1$ up to $0.034$ for $\lambda_2$ and $\delta_2$ with overall epistemic uncertainty of $0.029$ indicating an overall low model predictive uncertainty. Bias remains low with all parameter biases $<0.04$, indicating minimal systematic over or under parameter prediction. The low bias and overall average RMSE confirm that our model reliably recovers parameters from noisy time-series data with high accuracy. Figure \ref{fig:true_vs_pred_scatter} contrasts true and predicted means with $\pm 1\sigma$ error bars for some representative parameters $\lambda_2$, $\delta_2$, $\eta_2$ and $\iota_2$ which all indicate close alignment. The overall distribution of predicted parameters across test samples also exhibit a close match to the true parameter distribution as seen in Figure \ref{fig:bar_comparisons_param_dist}(b). In addition, Figure \ref{fig:bar_comparisons_param_dist}(a) visualizes how true parameter means differ from our predicted parameter means. These visual representations confirm our low prediction errors obtained.

\begin{table*}[cols=9,pos=h]
\caption{Results (Dropout rate $= 0.1$)}
\label{results_0.1}
\begin{tabular*}{\tblwidth}{@{} LLLLLLLLL@{} }
\toprule
& & & & & \multicolumn{2}{c}{\textbf{Pre-recalibration}} & \multicolumn{2}{c}{\textbf{Post-recalibration}}\\
\hline
\textbf{Parameter} & \textbf{True Means} & \textbf{Predicted Means} & $|\text{\textbf{Bias}}|$ & \textbf{RMSE} & \textbf{Avg. STD} & \textbf{Coverage $\pm1\sigma$} & \textbf{Avg. STD} & \textbf{Coverage $\pm1\sigma$}\\
\midrule
$\lambda_1$ & 0.3833 & 0.3823 & 0.0010 & 0.036 & 0.029 & 63.9\% & 0.035 & 70.1\%\\
$\lambda_2$ & 0.4008 & 0.3754 & 0.0255 & 0.058 & 0.034 & 48.6\% & 0.055 & 70.1\%\\
$\gamma_1$ & 0.2916 & 0.2879 & 0.0037 & 0.042 & 0.028 & 47.4\% & 0.042 & 67.9\%\\
$\gamma_2$ & 0.2856 & 0.2543 & 0.0313 & 0.067 & 0.030 & 34.3\% & 0.069 & 71.1\%\\
$\delta_1$ & 0.4089 & 0.4094 & 0.0039 & 0.035 & 0.030 & 60.4\% & 0.035 & 67.5\%\\
$\delta_2$ & 0.3944 & 0.3700 & 0.0244 & 0.055 & 0.034 & 46.6\% & 0.055 & 67.9\%\\
$\eta_1$ & 0.2869 & 0.2682 & 0.0186 & 0.044 & 0.026 & 42.8\% & 0.045 & 68.1\%\\
$\eta_2$ & 0.2860 & 0.2729 & 0.0130 & 0.061 & 0.024 & 40.2\% & 0.063 & 68.9\%\\
$\rho_1$ & 0.1416 & 0.1459 & 0.0044 & 0.014 & 0.018 & 80.9\% & 0.013 & 66.7\%\\
$\rho_2$ & 0.1419 & 0.1481 & 0.0063 & 0.023 & 0.023 & 67.3\% & 0.025 & 70.5\%\\
$\iota_1$ & 0.3329 & 0.3351 & 0.0022 & 0.060 & 0.032 & 32.7\% & 0.062 & 61.6\%\\
$\iota_2$ & 0.3193 & 0.3255 & 0.0062 & 0.017 & 0.028 & 85.7\% & 0.017 & 69.3\%\\
\hline
\textbf{Overall} & & & & \textbf{0.043} & \textbf{0.029} & & \textbf{0.043} & \\
\bottomrule
\end{tabular*}
\end{table*}

\subsection{Uncertainty Recalibration}
\begin{figure}
    \centering
    \includegraphics[width=0.9\columnwidth]{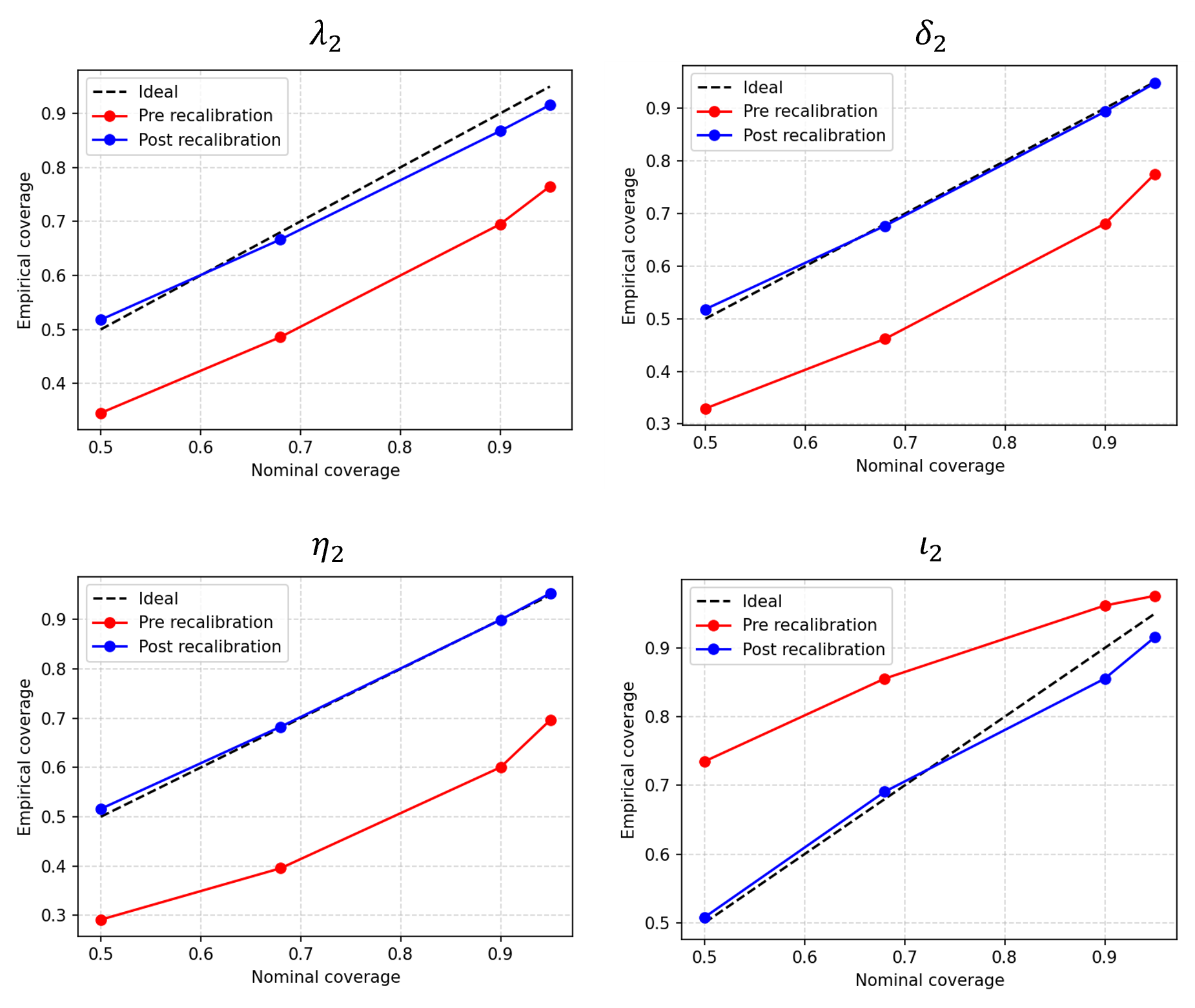}
    \caption{Uncertainty calibration curves for selected parameters. The red curves indicate pre-recalibration (coverage-based calibration) and blue curves represent post-recalibration (STD scaling).}
    \label{calibration_new}
\end{figure}

We assessed the calibration of the predicted uncertainties using coverage-based analysis at multiple nominal confidence levels $[0.5,~0.68,~0.9,~0.95]$. Prior to recalibration, our model exhibited systematic over-confidence and under-confidence in predicting the parameters. For example, only about $32.7\%$ of the true values fell within the predicted $\pm 1\sigma$ interval ($68\%$ confidence level) for $\iota_1$ indicating over-confidence, while for $\iota_2$, about $85.7\%$ of the true values fell within the predicted $\pm1\sigma$ interval, indicating very high model under-confidence leading to this over-coverage. Calibration curves (red curves) confirmed this, with most parameters showing empirical coverage well below the ideal diagonal as seen in Figures \ref{calibration_new} \& \ref{dp0.1_recals_others}. Thus, these reflected highly unreliable interval estimates.

After applying the STD scaling approach, our calibration improved substantially. As seen in Figures \ref{calibration_new} \& \ref{dp0.1_recals_others}, notice that the post-recalibration curves (in blue) moved closer to the ideal diagonal compared to the pre-recalibration curves (in red). For example, our model achieved a near perfect coverage at all confidence levels for $\delta_2$ and $\eta_1$ and specifically $67.9\%$ and $68.1\%$ at the $68\%$ nominal level (i.e. $\pm 1\sigma$) respectively. This very well indicates that predicted uncertainty intervals now better reflect actual error variability. However, some parameters (e.g. $\gamma_2$, $\rho_2$) showed slightly conservative behavior, with empirical coverage slightly exceeding the $68\%$ nominal level. This suggests that the model's predictive intervals may be marginally wider than necessary, favoring caution over precision in those cases.

In addition to these calibration plots, Table \ref{ence_results} reports the calibration error, quantified using the Expected Normalized Calibration Error (ENCE) metric. Before recalibration, our ENCE results showed relatively high percentages across all the parameters with some as high as 123.2\% for $\iota_1$ and 117.4\% for $\gamma_2$, consistent with the miscalibration observed in the coverage plots. Following STD scaling (post-recalibration), ENCE for $\iota_1$ and $\gamma_2$ significantly decreased by about 106\% and 107\% respectively. This significant decrease in the calibration error was consistent for all parameters after recalibration and indicates that a simple recalibration can reduce systematic over-confidence and under-confidence while maintaining the NN model's predictive accuracy.

\begin{figure*}
    \centering
    \includegraphics[width=0.8\linewidth]{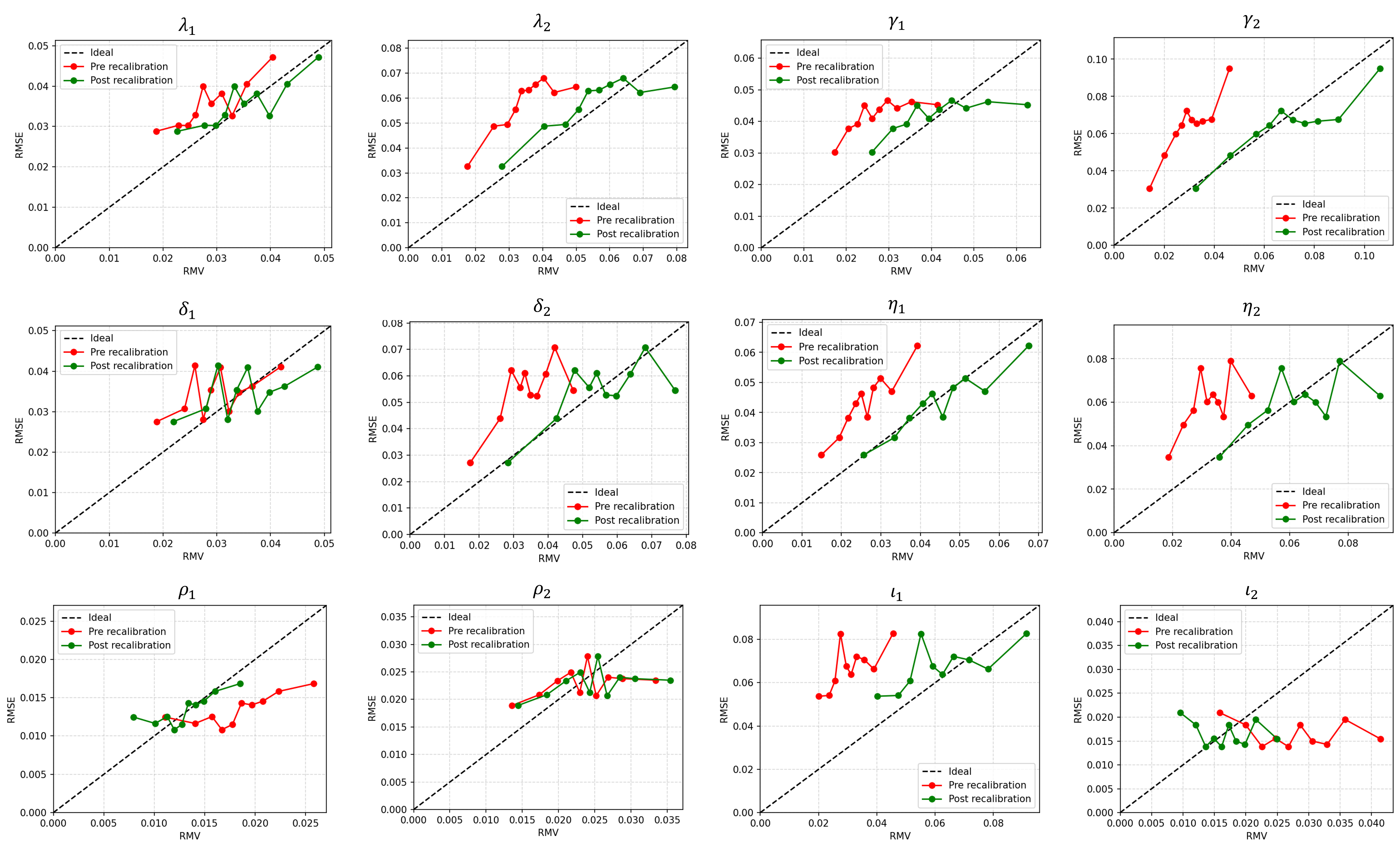}
    \caption{Pre- and Post-recalibration reliability plots for all predicted parameters}
    \label{fig:reliabs}
\end{figure*}

Figure \ref{fig:reliabs} shows a reliability plot of the RMSE against the RMV to further access and support calibration results. Before recalibration (red curves), most parameters exhibited systematic deviations from the diagonal, with RMSE exceeding RMV. This indicates that the model's predictive intervals were narrower than they should be, consistent with the coverage analysis showing over-confidence. However, the relationship between RMSE and RMV improved substantially after recalibration, with points shifting closer to the ideal diagonal.
This plot complements the coverage results by providing a bin-wise comparison between the predicted and empirical errors. The improvements observed post-recalibration confirm that the STD-scaling approach effectively corrected for systematic underestimation of uncertainty, leading to more trustworthy predictive intervals.

\begin{table*}[cols=4,pos=h]
\caption{Summary of Pre- and Post-recalibration Errors}
\label{ence_results}
\begin{tabular*}{\tblwidth}{@{} LLLL@{} }
\toprule
\textbf{Parameter} & \textbf{ENCE (Pre-recalibration)} & \textbf{ENCE (Post-recalibration)} & $\Delta$\textbf{ENCE}\\
\midrule
$\lambda_1$ & 26.3\% & \textbf{9.8\%} & 16.5\%\\
$\lambda_2$ & 69.5\% & \textbf{12.5\%} & 57.0\%\\
$\gamma_1$   & 56.7\% & \textbf{13.6\%} & 43.0\%\\
$\gamma_2$   & 117.4\% & \textbf{9.8\%} & 107.5\%\\
$\delta_1$   & 20.4\% & \textbf{16.7\%} & 3.7\%\\
$\delta_2$   & 62.9\% & \textbf{11.0\%} & 51.8\%\\
$\eta_1$     & 67.1\% & \textbf{5.9\%} & 61.2\%\\
$\eta_2$     & 88.0\% & \textbf{12.9\%} & 75.0\%\\
$\rho_1$     & 26.6\% & \textbf{11.8\%} & 14.7\%\\
$\rho_2$     & 18.6\% & \textbf{17.6\%} & 1.1\%\\
$\iota_1$    & 123.2\% & \textbf{17.0\%} & 106.2\%\\
$\iota_2$    & 41.6\% & \textbf{29.1\%} & 12.5\%\\
\bottomrule
\end{tabular*}
\end{table*}

Overall, our results demonstrate that parameter inference in Stochastic Petri Net models is both feasible and informative, even under biological realistic noise and partial observability. The conservative tendency in the interval calibration (post-recalibration) may be advantageous in public health contexts, where underestimating uncertainty can lead to overconfident decisions.

\subsection{Usefulness and Implications}
From a biological modeling perspective, these findings are important for three reasons:
\begin{enumerate}
    \item[i)] \textit{Petri Net Patch Modeling:} This provides a way to model complex patch dynamics of exogenous time-dependent environmental variables. The Petri Net framework implements a continuous time Markov process as shown in the SPIKE toolkit provides as reliable and robust simulation 
    \item[ii)] \textit{Trustworthy Uncertainty Estimates:} In downstream epidemiological simulations, having reliable uncertainty estimates is just as important as having accurate point predictions. Public health decisions often rely on the credibility of confidence or credible intervals, not just the mean prediction. Our results after recalibration ensured empirical values are closer to nominal targets, strengthening confidence that the predicted uncertainty bounds can be meaningfully interpreted in biological applications.
    \item[iii)] \textit{Parameter Heterogeneity:} The variability in calibration performance across parameters (some being over-confident, others under-confident) reflects differences in identifiability given the synthetic data. Parameters with weak or indirect influence on the Petri Net dynamics tended to have higher ENCE and worse coverage. This suggests that uncertainty calibration can also serve as a diagnostic tool to identify which biological parameters are harder to recover.
\end{enumerate}

\section{Conclusion and Future Work}
In this work, we proposed a neural-surrogate framework for recovering inherent parameters from covariate-dependent transition rates in Stochastic Petri Net (SPN) models, trained entirely on simulated token trajectories. Our approach avoids explicit likelihood evaluation and leverages a lightweight 1D-ResNet to learn mappings from partially observed dynamics. Despite missing event data of $10\%$, the surrogate achieved robust recovery with an RMSE of $0.043$, demonstrating that accurate inference is possible in partially observed systems with unavailable or intractable likelihoods. Monte Carlo dropout with STD scaling further enabled calibrated uncertainty estimates, making the method suitable for real-time or decision-critical settings.

This work serves as a first step toward fast, data-driven inference in SPNs with covariate-dependent dynamics. Future work will involve formal benchmarking against existing simulation-based methods such as Approximate Bayesian Computation (ABC) or ODE-based inference approaches where feasible, and testing the model’s generalization to more complex SPN topologies and real-world datasets.

\vspace{0.05in}
All of our codes for this work can be found here:\\
\begin{small}
    \url{https://github.com/BrightManu-lang/SPN-param-recovery.git}
\end{small}

\section*{Funding}
This work was supported by NIH grant 5R01GM131405-02 encompassing the efforts of Bright Manu and Trevor Reckell under primary investigators Petar Jevti\'{c} and Beckett Sterner.

\appendix
\section{Appendix}
\subsection{Supplementary Figures}
\begin{figure*}
    \centering
    \includegraphics[width=0.58\linewidth]{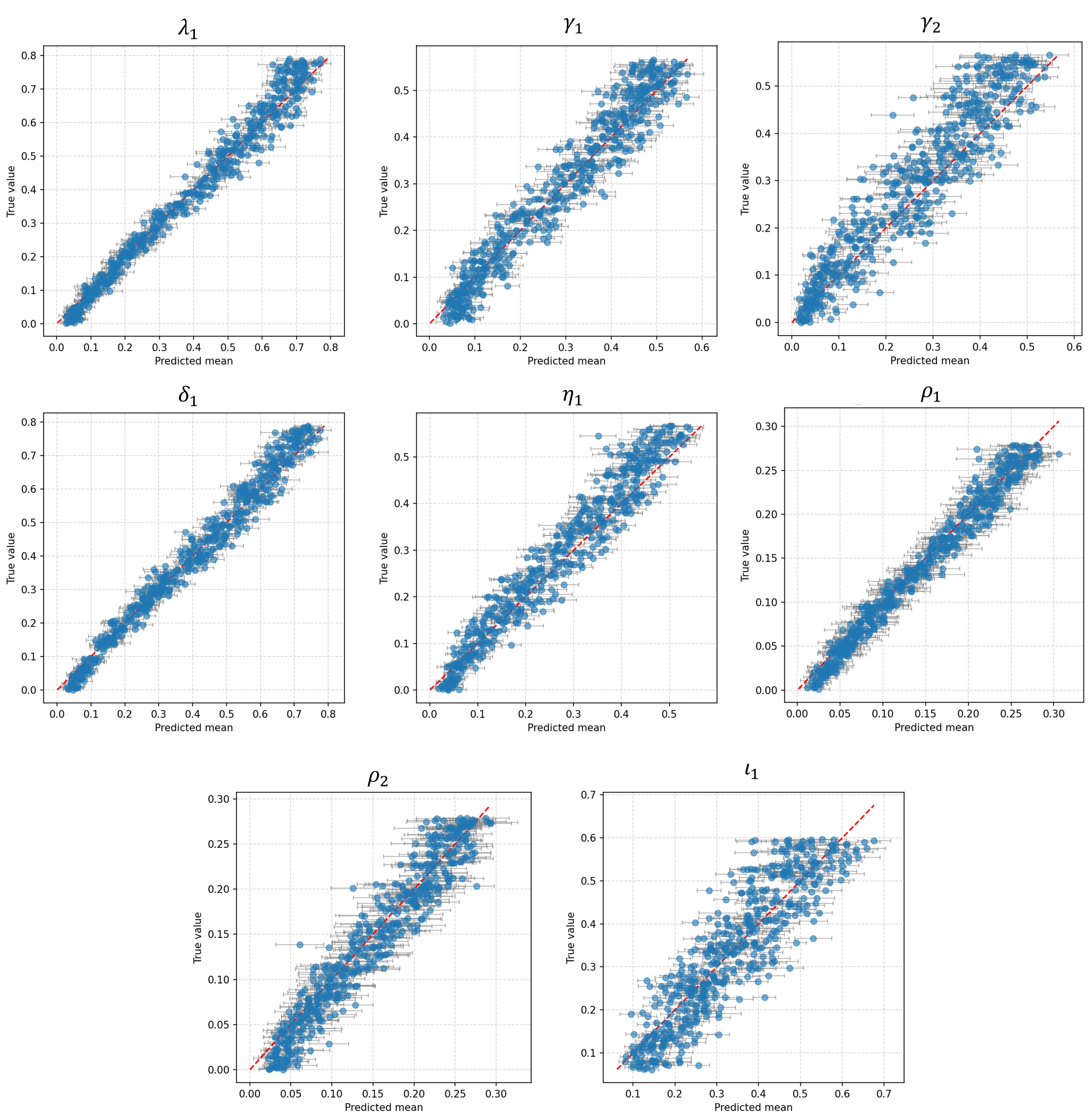}
    \caption{True versus Predicted sample parameters with $\pm1\sigma$ error bars}
    \label{dp0.1_true_vs_preds_others}
\end{figure*}

\begin{figure*}
    \centering
    \includegraphics[width=0.58\linewidth]{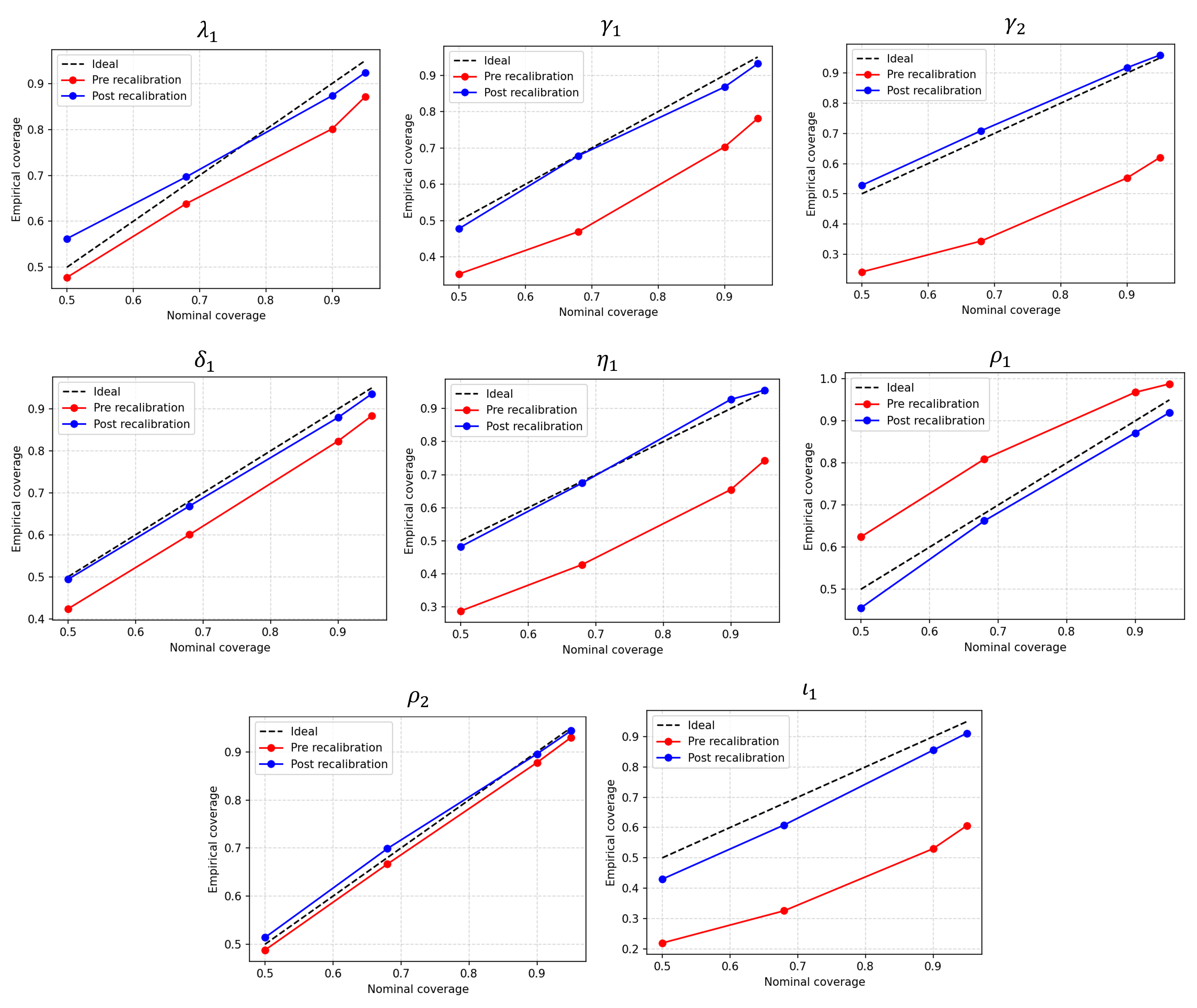}
    \caption{Pre- and Post-recalibration plots for the other predicted parameters}
    \label{dp0.1_recals_others}
\end{figure*}

First we show the true and predicted sample parameter plots (Figure \ref{dp0.1_true_vs_preds_others}) as well as the calibration plots for the other parameters (Figure \ref{dp0.1_recals_others}) not shown in the main results.

\subsection{Dropout Rate $0.2$}
Following the suggested dropout rates in the work of \cite{gal2016dropout}, we also run our model with a dropout rate of $0.2$. We note that we used the same dropout rate in training our model as the same dropout rate used during inference time (One can set up the NN model without a dropout rate or tune the dropout rate for the purpose of model performance and use a different dropout rate during inference time). The results are displayed in Table \ref{results_dp0.2}.

\begin{table*}[cols=9,pos=h]
\caption{Results (Dropout rate $= 0.2$)}
\label{results_dp0.2}
\begin{tabular*}{\tblwidth}{@{} LLLLLLLLL@{} }
\toprule
& & & & & \multicolumn{2}{c}{\textbf{Pre-recalibration}} & \multicolumn{2}{c}{\textbf{Post-recalibration}}\\
\hline
\textbf{Parameter} & \textbf{True Means} & \textbf{Predicted Means} & $|\text{\textbf{Bias}}|$ & \textbf{RMSE} & \textbf{Avg. STD} & \textbf{Coverage $\pm1\sigma$} & \textbf{Avg. STD} & \textbf{Coverage $\pm1\sigma$}\\
\midrule
$\lambda_1$ & 0.3833 & 0.3750 & 0.0083 & $0.037$ & $0.043$ & 77.3\% & $0.039$ & 72.7\%\\
$\lambda_2$ & 0.4008 & 0.3896 & 0.0112 & $0.060$ & $0.045$ & 58.6\% & $0.054$ & 66.7\%\\
$\gamma_1$ & 0.2916 & 0.2877 & 0.0039 & $0.045$ & $0.037$ & 58.0\% & $0.045$ & 65.5\%\\
$\gamma_2$ & 0.2856 & 0.2774 & 0.0082 & $0.065$ & $0.040$ & 48.4\% & $0.064$ & 69.5\%\\
$\delta_1$ & 0.4089 & 0.4031 & 0.0058 & $0.036$ & $0.044$ & 75.5\% & $0.036$ & 68.7\%\\
$\delta_2$ & 0.3944 & 0.3716 & 0.0228 & $0.059$ & $0.045$ & 55.4\% & $0.060$ & 69.7\%\\
$\eta_1$ & 0.2868 & 0.2698 & 0.0171 & $0.046$ & $0.036$ & 57.8\% & $0.046$ & 67.5\%\\
$\eta_2$ & 0.2860 & 0.2764 & 0.0097 & $0.066$ & $0.042$ & 47.2\% & $0.064$ & 66.7\%\\
$\rho_1$ & 0.1416 & 0.1391 & 0.0025 & $0.018$ & $0.025$ & 84.3\% & $0.017$ & 67.9\%\\
$\rho_2$ & 0.1419 & 0.1445 & 0.0027 & $0.028$ & $0.028$ & 63.1\% & $0.027$ & 61.2\%\\
$\iota_1$ & 0.3330 & 0.3285 & 0.0045 & $0.071$ & $0.041$ & 41.2\% & $0.069$ & 66.9\%\\
$\iota_2$ & 0.3193 & 0.3188 & 0.0005 & $0.022$ & $0.038$ & 85.7\% & $0.021$ & 61.8\%\\
\hline
\textbf{Overall} & & & & $\textbf{0.046}$ & $\textbf{0.039}$ & & $\textbf{0.045}$  & \\
\bottomrule
\end{tabular*}
\end{table*}

\begin{figure*}
    \centering
    \includegraphics[width=0.8\linewidth]{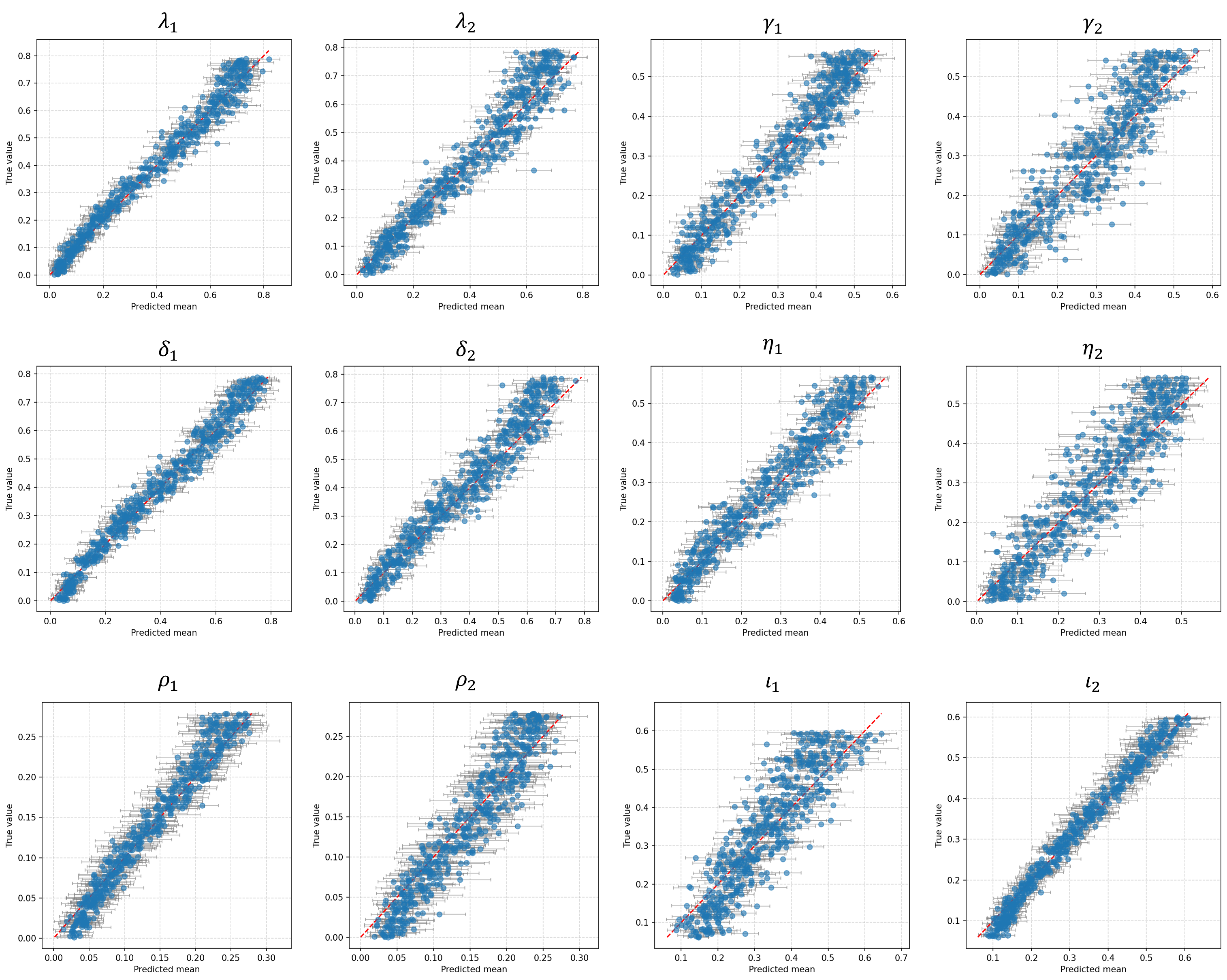}
    \caption{True versus Predicted sample parameters with $\pm1\sigma$ error bars (0.2 dropout)}
    \label{fig:placeholder}
\end{figure*}

\begin{figure*}
    \centering
    \includegraphics[width=0.8\linewidth]{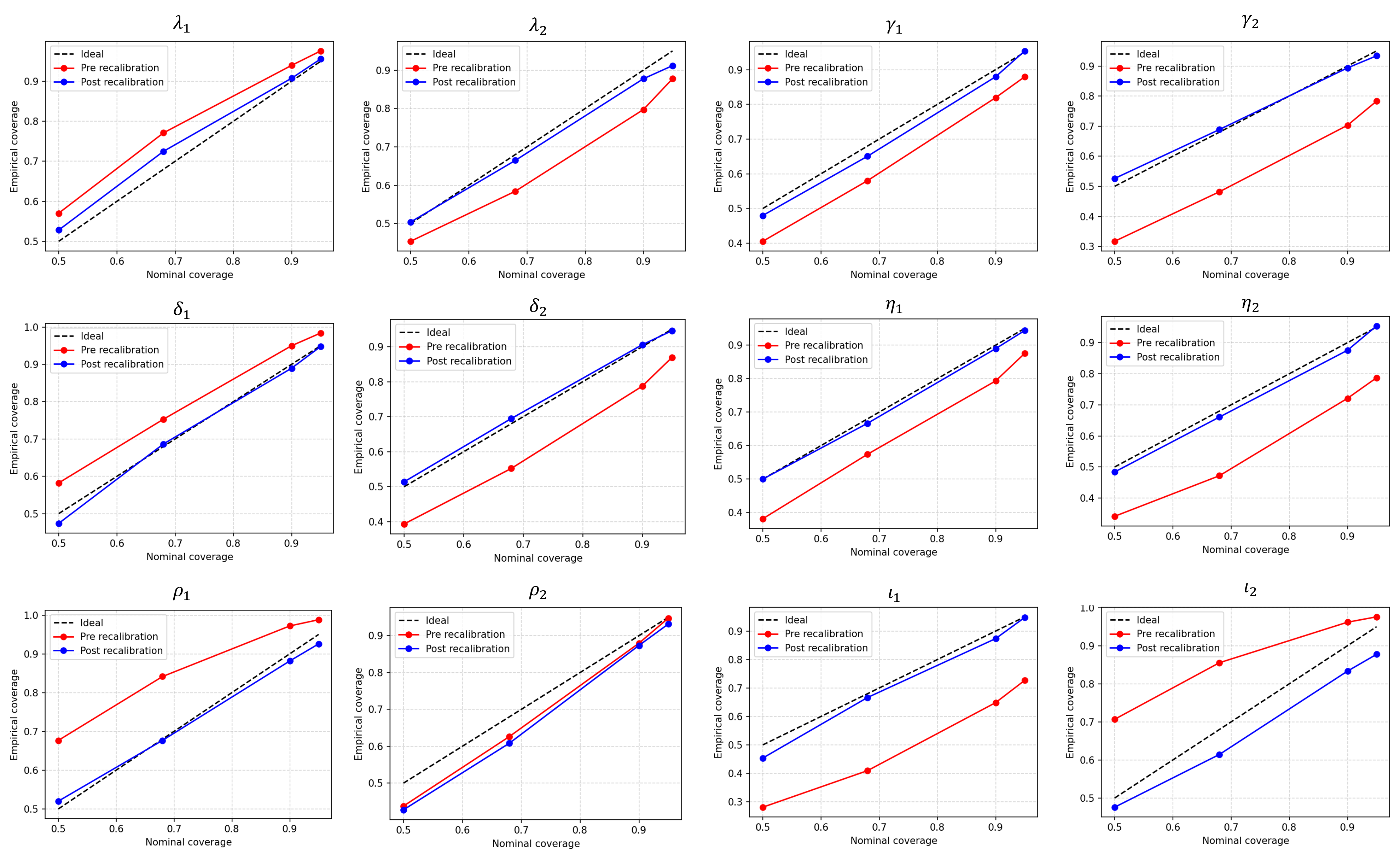}
    \caption{Pre- and Post-recalibration plots for all predicted parameters (0.2 dropout)}
    \label{fig:placeholder}
\end{figure*}

\begin{figure*}
    \centering
    \includegraphics[width=0.8\linewidth]{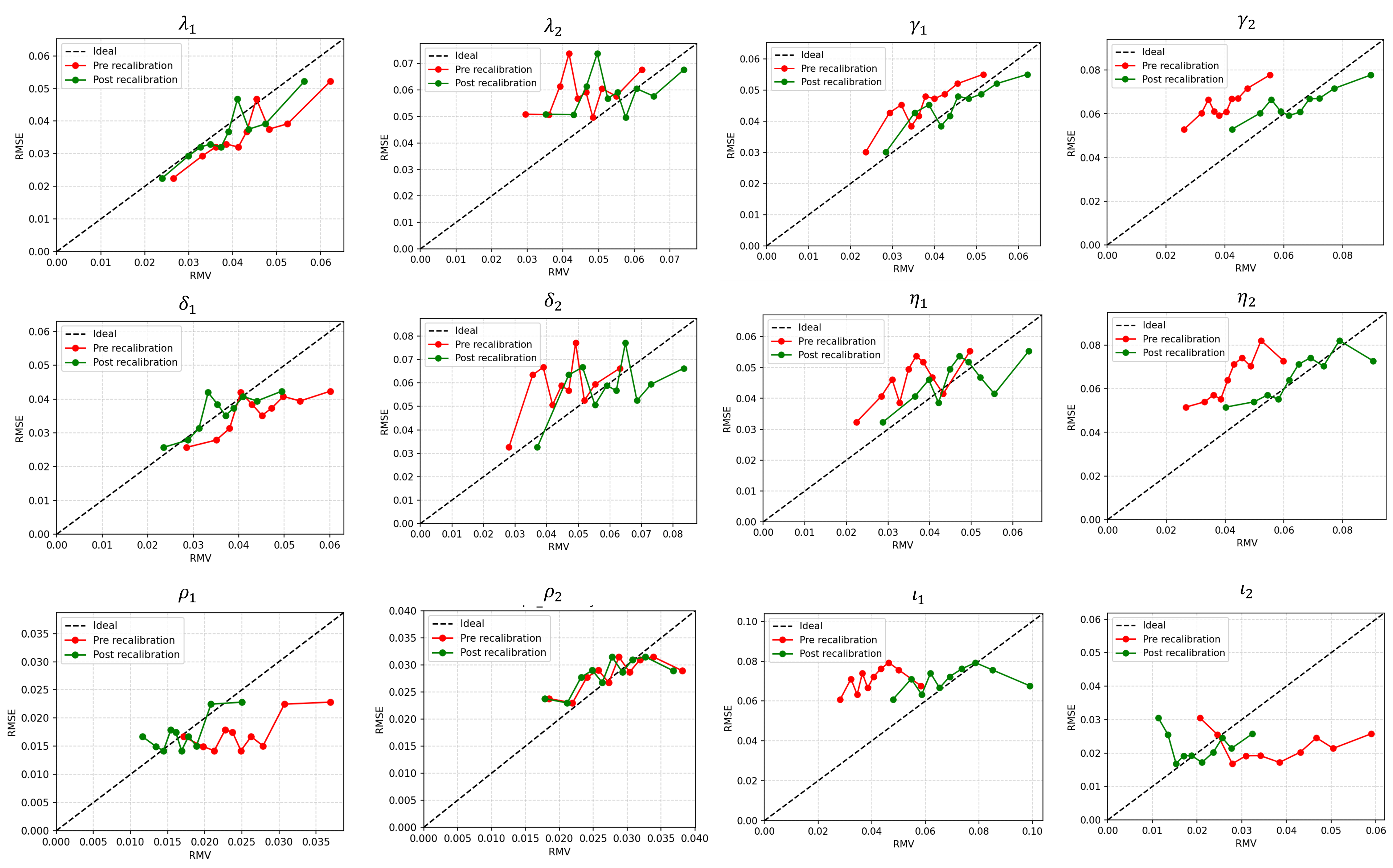}
    \caption{Pre- and Post-recalibration reliability plots for all predicted parameters for 50 stochastic forward passes}
    \label{fig:mc_calibs}
\end{figure*}

\subsection{Hyperparameter Optimization}
The MC dropout algorithm has tunable parameter (dropout rate) which potentially affects performance. The community often adopts 0.1 and 0.2 as defaults and we investigated whether there is a better option. We conducted a hyperparameter sweep using the Optuna algorithm. The sweep targeted key parameters influencing model behavior, generalization and most importantly uncertainty. Parameters explored include Learning rate: $[1e-4, 3e-2]$, batch size: $\{8, 16, 32\}$ and dropout rate: $[0.0, 0.5]$ as seen in Table \ref{hyperparams}.

This was carried out on the validation set for 10 different trials, each trial having different set of combinations of hyperparameters, and the optimal configuration was selected based on the lowest validation loss. The best hyperparameters were as follows

\begin{table}[width=.9\linewidth, cols=3,pos=h]
\caption{Tuned Hyperparameters}
\label{hyperparams}
\begin{tabular*}{\tblwidth}{@{} LLL@{} }
\toprule
\textbf{Hyperparameter} & \textbf{Range/Values} & \textbf{Tuned/Best} \\
\midrule
Dropout rate & $[0.0,0.5]$ & $0.08$ \\
Learning rate & $[1e-4,3e-2]$ & $7.6e-4$ \\
Batch size & $\{8, 16, 32\}$ & $16$ \\
\bottomrule
\end{tabular*}
\end{table}

At inference time, we explored multiple stochastic forward passes for the following Monte Carlo samples $\{10, 100, \\250, 500\}$. As depicted in Figure \ref{fig:mc_calibs}, our tuned dropout rate used at inference time did not show much improvement even after trying multiple stochastic forward passes.

\begin{figure*}
    \centering
    \includegraphics[width=0.9\linewidth]{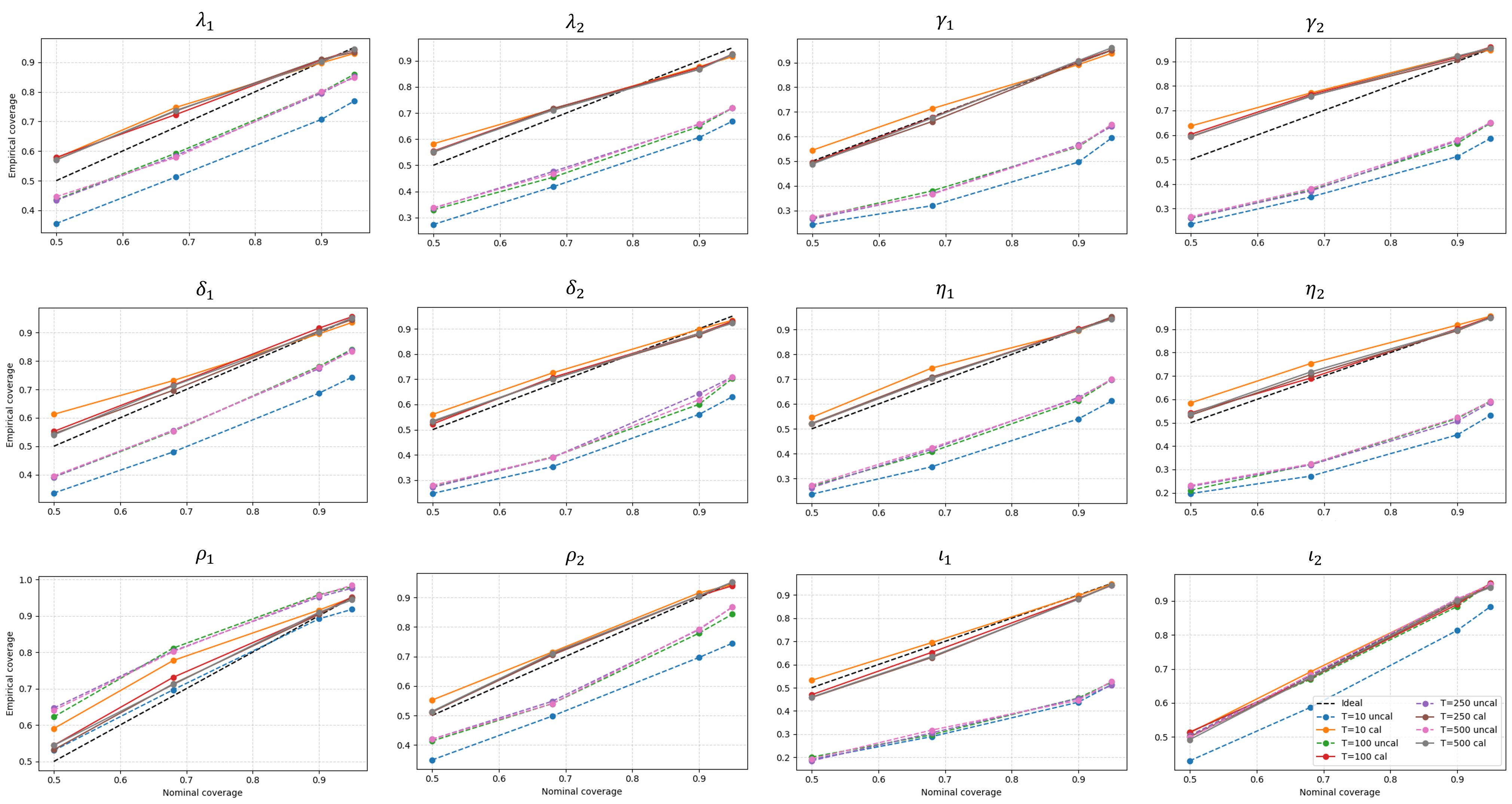}
    \caption{Pre- and Post-recalibration plots for all predicted parameters for Multiple Monte Carlo samples (0.08 dropout)}
    \label{fig:mc_calibs}
\end{figure*}


\bibliographystyle{cas-model2-names}

\bibliography{refs}





\end{document}